\documentclass[letterpaper]{article} 
\pdfoutput=1
\usepackage{aaai24}  
\usepackage{times}  
\usepackage{helvet}  
\usepackage{courier}  
\usepackage[hyphens]{url}  
\usepackage{graphicx} 
\usepackage{natbib}  
\usepackage{caption} 
\usepackage{algorithm}
\usepackage{algorithmic}
\usepackage{graphicx}
\usepackage{amsmath}
\usepackage{amssymb}
\usepackage{booktabs}
\usepackage{multirow}
\usepackage{amsmath,amsfonts}
\usepackage{algorithmic}
\usepackage{algorithm}
\usepackage{array}
\usepackage[caption=false,font=normalsize,labelfont=sf,textfont=sf]{subfig}
\usepackage{textcomp}
\usepackage{url}
\usepackage{verbatim}
\usepackage{graphicx}
\usepackage{cite}
\usepackage{graphicx}
\usepackage{booktabs}
\usepackage{times}
\usepackage{algorithm}
\usepackage{algorithmic}
\usepackage{graphicx}
\usepackage{epstopdf}
\usepackage{verbatim}
\usepackage{epsfig}
\usepackage{amsmath}
\usepackage{amssymb}
\usepackage{color}
\usepackage{booktabs}
\usepackage{xcolor}
\usepackage{color, colortbl}
\usepackage{makecell}

\usepackage{fancyhdr}
\usepackage{lipsum} 
\usepackage{newfloat}
\usepackage{listings}
\DeclareCaptionStyle{ruled}{labelfont=normalfont,labelsep=colon,strut=off} 
\floatstyle{ruled}
\newfloat{listing}{tb}{lst}{}
\floatname{listing}{Listing}
\nocopyright

\title{ALPS: An Auto-Labeling and Pre-training Scheme for Remote Sensing Segmentation With Segment Anything Model}
\author{
    Song Zhang \equalcontrib \textsuperscript{\rm 1} 
    Qingzhong Wang \equalcontrib \textsuperscript{\rm 2},
    Junyi Liu \textsuperscript{\rm 1}, 
    Haoyi Xiong\thanks{Corresponding author} \textsuperscript{\rm 2},
}
\affiliations{
    \textsuperscript{\rm 1}Aerospace Information Research Institute, Chinese Academy of Sciences. 
    \textsuperscript{\rm 2}Baidu Inc.\\
}

\fancypagestyle{firstpage}{
  \fancyhf{} 
  \fancyfoot[L]{\small{This work has been submitted to the IEEE for possible publication. Copyright may be transferred without notice, after which this version may no longer be accessible.}}
}

\usepackage{bibentry}

\begin{document}

\maketitle

\thispagestyle{firstpage}

\begin{abstract}
In the fast-growing field of Remote Sensing (RS) image analysis, the gap between massive unlabeled datasets and the ability to fully utilize these datasets for advanced RS analytics presents a significant challenge. To fill the gap, our work introduces an innovative auto-labeling framework named ALPS (Automatic Labeling for Pre-training in Segmentation), leveraging the Segment Anything Model (SAM) to predict precise pseudo-labels for RS images without necessitating prior annotations or additional prompts. The proposed pipeline significantly reduces the labor and resource demands traditionally associated with annotating RS datasets. By constructing two comprehensive pseudo-labeled RS datasets via ALPS for pre-training purposes, our approach enhances the performance of downstream tasks across various benchmarks, including iSAID and ISPRS Potsdam. Experiments demonstrate the effectiveness of our framework, showcasing its ability to generalize well across multiple tasks even under the scarcity of extensively annotated datasets, offering a scalable solution to automatic segmentation and annotation challenges in the field. In addition, the proposed a pipeline is flexible and can be applied to medical image segmentation, remarkably boosting the performance. Note that ALPS utilizes pre-trained SAM to semi-automatically annotate RS images without additional manual annotations. Though every component in the pipeline has bee well explored, integrating clustering algorithms with SAM and novel pseudo-label alignment significantly enhances RS segmentation, as an off-the-shelf tool for pre-training data preparation. Our source code is available at: https://github.com/StriveZs/ALPS.
\end{abstract}

\section{Introduction}
\label{sec:intro}

Semantic segmentation represents a critical task within computer vision, involving pixel-level classification that plays a pivotal role across various applications. This technique, essential for attributing distinct labels to each pixel, facilitates a comprehensive understanding of spatial relationships within images. It stands as a cornerstone in applications ranging from lane detection in autonomous driving \cite{b1} to UAV geolocalization efforts \cite{b2}. Distinguished from image prediction, semantic segmentation delivers an in-depth object delineation, integrating valuable spatial context. This technique has been widely used in multiple areas, especially in addressing complex, data-intensive challenges inherent in remote sensing (RS) \cite{b3}. In the RS domain, the applications of semantic segmentation span environmental surveillance \cite{b4, b5}, agricultural crop cover and variety determination \cite{b6, b7, b8}, forestry for species recognition \cite{b9}, to urban planning through building classification and land-use analysis \cite{b10, b11, b12}.  


\begin{figure} [!t]
    \centering 
    \includegraphics[width=1\linewidth]{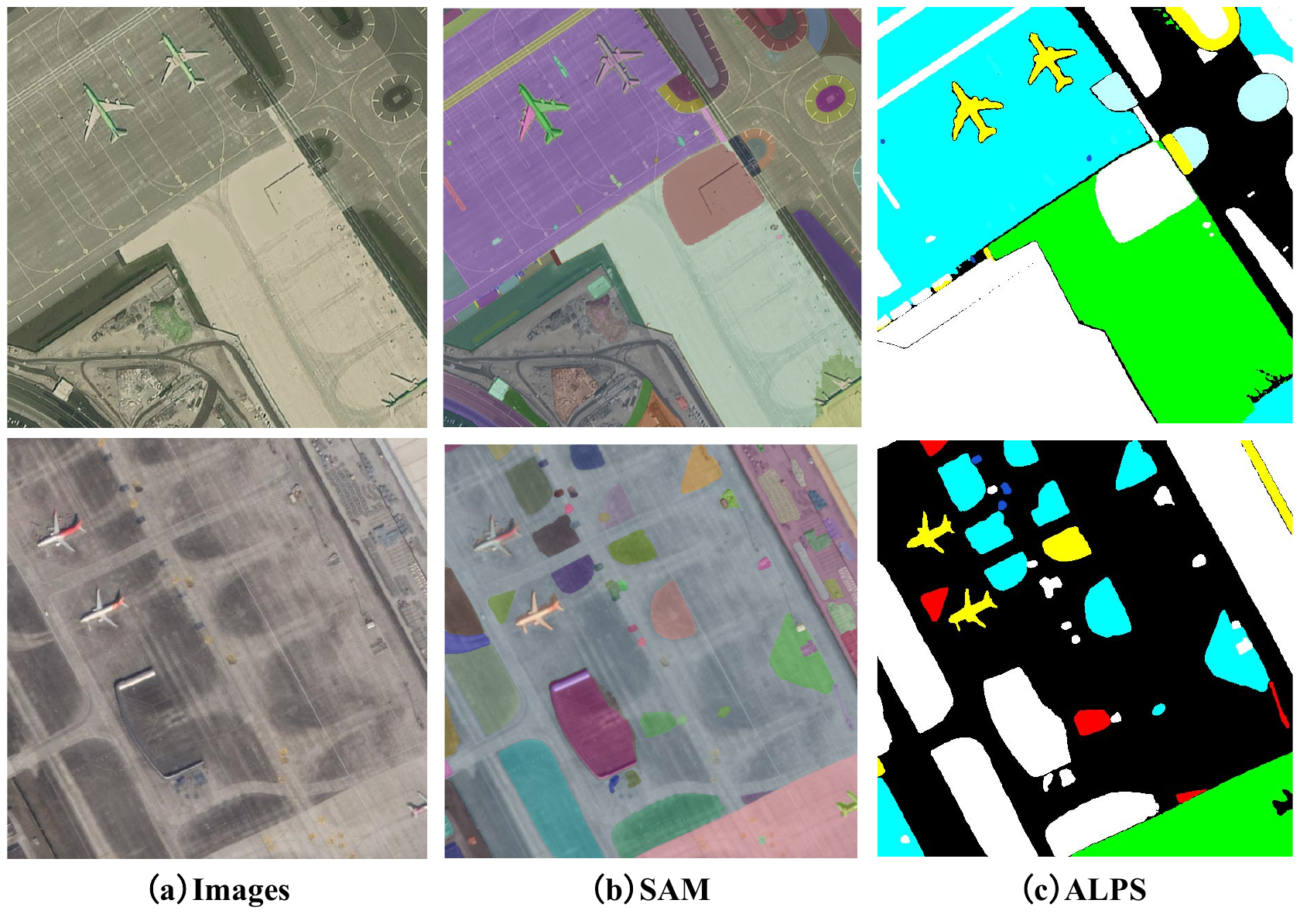}
    \caption{Some examples of SAM segmentation results and ALPS segmentation results on remote sensing images. (a)~some remote sensing images obtained from the SAMRS \cite{samrs}, (b)~segmentation results predicted by the SAM \cite{sam} without any prompts, (c)~semantic segmentation results generated by our ALPS without any prompts.} 
    \label{fig:teaser_figure}
\end{figure}

The advancement of earth observation technologies has led to the generation of abundant RS images. Yet, annotating these images with precise, pixel-level semantic labels remains laborious and expensive, making the acquisition of extensively annotated datasets for downstream supervised segmentation tasks challenging. In response, researchers have turned to weakly-supervised techniques that employ minimal forms of supervision, such as scribbles \cite{b13,b14,b15,b16,b17}, bounding boxes \cite{b18,b19,b20,b21}, clicks \cite{b22}, and image-level tags \cite{b20, b15, b21}, or semi-supervised methods that label only a subset of the dataset \cite{b18,b20,b23,b24,b25}, both strategies significantly reducing the need for exhaustive human annotation. Despite advancements, the crux of semantic segmentation research still necessitates some degree of manual labeling for training neural networks. Consequently, a new wave of weakly semantic segmentation approaches has emerged, eschewing labels altogether. These pioneering methods harness pixel-level self-supervised representation learning, introducing techniques such as cross-view consistency \cite{b26,b27}, edge detection \cite{b28,b29}, and saliency prior \cite{b30} to autonomously parse and understand RS imagery.


On the other hand, the Segment Anything Model (SAM) \cite{sam} has emerged as a groundbreaking advancement in computer vision, primarily for its outstanding ability in object segmentation. The robust generalization capability enables SAM to exhibit exceptional zero-shot segmentation, even in specialized domains like remote sensing (RS) images \cite{b31,b32,b33} and medical imagery \cite{b34,b35,b36}, leveraging training from a comprehensive natural image dataset. Extensions of SAM, such as SAMFeat \cite{samfeat}, SA3D \cite{sa3d}, and SAM-Track \cite{sam-track}, further amplify its utility in tasks ranging from local feature learning to 3D segmentation and video object tracking. Uniquely, in the RS domain, SAM showcases profound zero-shot segmentation potential. Studies like Ren \cite{ren} and Julka \cite{julka} have validated SAM’s versatility across different datasets and applications, including planetary geography. New methods like Text2Seg \cite{text2seg} and RSPrompter \cite{rsprompter} have extended the applications of SAM with text-based guidance and novel prompt learning methods for improved semantic segmentation and instance segmentation, respectively. Moreover, leveraging SAM for data annotation has been explored, with SAMRS \cite{samrs} demonstrating its efficacy in generating large-scale, accurately labeled datasets, showcasing its  efficiency and effectiveness of data annotation across computer vision tasks.

However, leveraging SAM to automatically annotate RS images for semantic segmentation presents significant challenges:
\begin{itemize}
    \item \emph{Detection Annotations and Additional Prompts:} Existing solutions based on SAM, such as SAMRS \cite{samrs}, usually need the annotations of object detection or external prompts for semantic label predictions from RS images. However, the objective of our work is to bypass these constraints, enabling SAM to autonomously generate labels across diverse RS scenarios without external prompts.

    \item \emph{Random Color Label Assignments by SAM:} When used without any prompts, SAM segments all elements within an image and assigns random color labels. These color labels for objects of the same or different categories lack consistency, rendering them unsuitable for direct use in downstream segmentation training tasks.

    \item \emph{Inconsistency in Segmentation Results:} The segmentation outcomes for the same image by vanilla SAM can vary. For example, in Figure~\ref{fig:teaser_figure}, SAM sometimes identifies the wings and fuselage of a plane as distinct parts, and at other times, considers them as a single entity. This randomness adds another layer of complexity in developing a unified strategy for automatic labeling.
\end{itemize}
Our study focuses on overcoming these hurdles by formulating a method that allows SAM to allocate unified pixel-level labels to objects of the same category, thereby facilitating an auto-labeling scheme for massive unlabeled data.  

To tackle above challenges, this work introduces \textbf{ALPS} (Automatic Labeling for Pre-training in Segmentation), a novel auto-labeling framework that leverages the capabilities of the vanilla Segment Anything Model (SAM) for annotating vast quantities of unlabeled remote sensing (RS) images for semantic segmentation. Utilizing the vanilla SAM, we first generate \emph{Unlabeled Instance Mask sets} (UiMs) for each image using the raw output of SAM, then identifying unique segmentation outlines without associated labels. Through an innovative process, high-level features of each mask are extracted to define a \emph{pseudo feature label}~(PFL), subsequently clustered via online K-means \cite{k-means} to assign a \emph{pseudo class label}~(PCL) to each mask. This approach automates the construction of pre-training datasets aimed at enhancing the performance of downstream RS segmentation tasks. Though every component in the proposed pipeline is a well-explored method, their specific applications in conjunction with SAM for RS images and the novel mechanism for aligning pseudo-labels represents a noteworthy adaptation and refinement for RS contexts.

Our technical contributions can be summarized as follows:
\begin{itemize}
    \item We systematically study the problem of efficiently annotating RS images for segmentation without manual prompts, leveraging a label-level distillation process from SAM. To best of our knowledge, this work is the first to enhance RS pre-training through distilling SAM on unlabeled datasets, by addressing the technical issues including lack of detection annotations, uses of additional prompts, and random \& inconsistent outputs of SAM on RS imaginary.
    
    \item We propose an effective and innovative auto-labeling framework, ALPS, which employs high-level feature clustering to achieve efficient labeling of large-scale datasets using the vanilla SAM. Note that the proposed framework leverages the pre-trained SAM, which is originally trained on human-annotated data. However, in our approach, no additional manual annotations are required for RS images, thus enabling a semi-automated annotation process without additional human intervention.

    \item Our extensive experiments on pseudo-labeled pre-training datasets, including iSAID \cite{isaid} and SAMRS \cite{samrs}, validate our methodology. By conducting segmentation pre-training, our studies underscore the effectiveness of leveraging vast RS segmentation data to mitigate task discrepancies and address the challenges posed by limited training data availability. Experiments show that our ALPS improves mIoU by up to 9.98\% compared to SAMRS*.

    \item The proposed is flexible. We also conduct experiments on medical image segmentation. The experimental results show that the proposed method notably improve the mIoU by at most 6.12\%.
\end{itemize} 
This approach has made a significant advancement in utilizing auto-labeled datasets for enhancing the accuracy and efficiency of segmentation models in remote sensing applications.

\section{Preliminaries}
In this section, we review the foundation model~(SAM) and remote sensing~(RS) datasets used in our framework. We primarily employ the vanilla SAM \cite{sam}, iSAID dataset \cite{isaid} and SAMRS dataset \cite{samrs}.

\subsection{Segment Anything Model}
Recently, the SAM \cite{sam}, a segmentation model, was recently introduced by Meta AI Research alongside the largest, most comprehensive segmentation dataset to date, SA-1B, containing 1.1 billion masks and 11 million images. 
Through specially designed training methods and a large-scale training data SA-1B, it offers not only support for interactive segmentation methods but also delivers outstanding zero-shot performance on a wide range of segmentation tasks. These two crucial features significantly enhance the applicability of SAM and make it a promising solution for various computer vision applications. Therefore, in this study, we adopt the vanilla SAM to predict the UiMs and utilize the encoder of vanilla SAM to extract the PFL for each mask.


\subsection{Datasets For Automatic Labeling}

\textbf{iSAID}: iSAID \cite{isaid} is a large-scale aerial image segmentation benchmark, which consists of 2,806 high resolution remote sensing~(RS) images. These images were collected from multiple sensors and platforms with multiple resolutions. The size of the original images ranges from 800 $\times$ 800 to 4000 $\times$ 13000. The iSAID dataset provides 655,451 instance annotations over 15 categories, which is one of the largest datasets for instance segmentation in remote sensing. The training set contains 1,411 images, the validation (val) set contains 458 images and the test set has 937 images. In this work, we only use the original images for our ALPS framework.

\textbf{SAMRS}: The SAMRS \cite{samrs} dataset, which stands for Segment Anything Model annotated RS segmentation dataset, is a large-scale segmentation dataset constructed using SAM and off-the-shelf object detection models trained on RS dataset, e.g. HRSC2016 \cite{HRSC2016}, DOTA-V2.0 \cite{DOTA}, DIOR \cite{DIOR}, and FAIR1M-2.0 \cite{FAIR1M}. This dataset consists of three subsets: SOTA, SIOR, and FAST. SOTA and SIOR are segmentation datasets containing common object categories, while FAST is a dataset specifically targeting diverse vehicles and grounds. The training set contains 8,685 images, and validation (val) set contains 4,667 images. In this study, we only use the original images for our ALPS framework.

\begin{figure*} [!t]
    \centering 
    \includegraphics[width=1\linewidth]{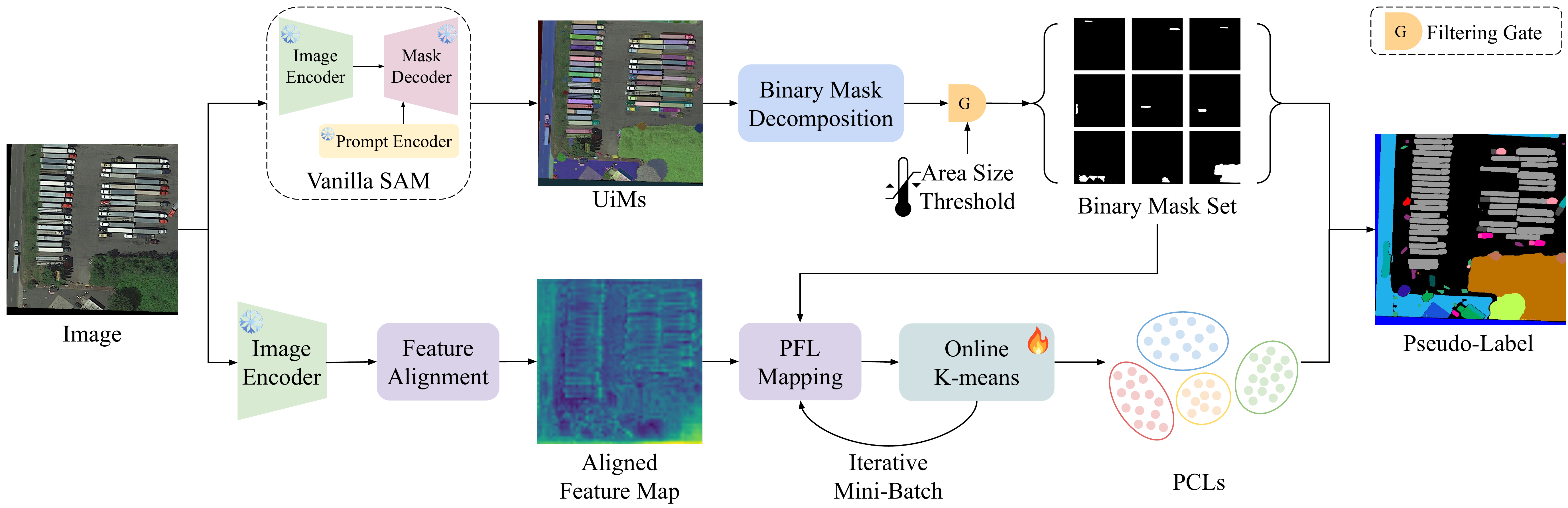}
    \caption{The illustration of our ALPS framework. Our mainly framework consists of two parts, which can obtain the binary mask set and PCL for each mask respectively.} 
    \label{fig:framework}
\end{figure*}

\section{Methodology}
This section presents the methodology of ALPS (Automatic Labeling for Pre-training in Segmentation).
\subsection{Pipeline}
Our ALPS framework is in light of achieving unsupervised automatic labeling on massive RS unlabeled data, meanwhile, the data with pseudo labels can be used to pre-train segmentation models to improve the performance of the downstream segmentation tasks. 
To achieve this goal, we introduce ALPS, an innovative auto-labeling framework based on the vanilla SAM, which comprises two parts -- Binary Mask Prediction and  Mask Class Association.
We first use SAM to obtain the binary masks of instances and then employ a clustering approach to assign a label to each mask. There three steps to achieve the smantic labels. Firstly, we obtain the \emph{pseudo feature label} (PFL) by extracting the regional feature of each mask. Secondly, we apply online K-means for clustering and finally, we assign a \emph{pseudo class label} (PCL) to each mask based on the clustering centers.
Fig \ref{fig:framework} illustrates the architecture of our ALPS framework.

Specifically, assume there is a large amount of unlabeled RS images $\left\{I_{i} \right\}_{i=0}^{N}$, given an input image $I_{i}$, our ALPS framework firstly uses the vanilla SAM without any additional prompts to obtain the \emph{Unlabeled Instance Mask set}~(UiMs) $\upsilon$, then we introduce a binary mask decomposition module to obtain the binary mask of each instance and design a filtering gate to filter out the abnormal masks to alleviate the erosion of the reliable pseudo-labels by the overly large background.

Furthermore, to assign the appropriate label to each mask, we employ the encoder of vanilla SAM \cite{sam} to extract the semantic features $F_{i}$ from the input image. Subsequently, these features are upsampled to match the scale of the binary mask, thereby facilitating feature alignment. Then we obtain the binary masks sequentially from the set and carry out PFL mapping in conjunction with the aligned feature map. The results of mapping are then fed into an online K-means \cite{k-means} to achieve iterative clustering to obtain the PCL for each mask. Finally, by integrating the outputs from the above two parts, we can generate the pseudo-labels through unsupervised automatic labeling for original images. Moreover, benefiting from the generalization of SAM and scalability of K-means, our ALPS framework can easily apply to any similar domain data.

\subsection{Binary Mask Prediction}
\label{sec:3.2}
\noindent \textbf{UiMs Prediction}: In our ALPS framework, we employ the officially provided, pre-trained ViT-H \cite{vit} based SAM model as the toolkit for UiMs prediction. We have leveraged the mechanism of automatic mask generation of vanilla SAM. Specific ally, we take an RS image $I_{i}$ and default a set of dense prompts  $\mathcal{P}$ as input, where the dense prompts are the default configuration of SAM, and outputs the corresponding UiMs in the form of a bitmap, the formulation is defined as Eq. \ref{eq:equation_2}:

\begin{equation}
 \label{eq:equation_2}
\begin{aligned}
    \text{UiMs}=S(I_{i}, \mathcal{P})
\end{aligned}
 \end{equation}

\noindent where $S(\cdot)$ represents the inference phase of vanilla SAM. 

\noindent \textbf{Binary Mask Generation}: After obtain the UiMs, we design a binary mask decomposition module and filtering gate to generate the binary mask set for later process. Specifically, we have developed a module to individually extract each mask from UiMs and convert it into a binary mask for separate storage, by leveraging the characteristic that each mask $M$ in UiMs has color differences in RGB values. Then, to alleviate the erosion of the instance mask by the overly large background mask, we designed a filtering gate $G(\cdot)$ to filter out the overly large background mask. First, we defined a instance proportion $\mathcal{PI}(\cdot)$, which is obtained by calculating the percentage of pixels occupied by the instance in the overall mask, defined as Eq. \ref{eq:equation_3}:

\begin{equation}
 \label{eq:equation_3}
\begin{aligned}
    \mathcal{PI}(M_{i})=\frac{num(p^{0}_{i})}{num(p^{0}_{i} + p^{255}_{i})}
\end{aligned}
 \end{equation}

\noindent where $num(\cdot)$ represents the number of pixels, $p^{0}_{i}$ represents the pixels occupied by the instance, where the RGB value of the instance in the binary mask is $(0,0,0)$, and $p^{255}_{i}$ represents the pixel occupied by the background.
Then we use an area size threshold $\sigma$ (default is 0.3) to filter out binary masks with too large instance proportions, thereby implementing the mechanism of the filtering gate $G(\cdot)$, defined as Eq. \ref{eq:equation_4}:

\begin{equation}
 \label{eq:equation_4}
    G(M_{i}) =  
    \begin{cases}
     M_{i} & if \; \; \mathcal{PI}(M_{i}) <= \sigma\\
     - & if \; \; \mathcal{PI}(M_{i}) > \sigma \\
    \end{cases}
 \end{equation}

\noindent where ``$-$'' indicates that this binary mask is discarded.

\subsection{Mask Class Association}
\label{sec:3.3}
\noindent \textbf{Feature Extraction and Alignment}: After filtering, the binary mask lacks inherent class information, making direct categorization a challenging task. Currently, in both machine learning and deep learning paradigms, features play a pivotal role by providing essential semantic information that facilitates the execution of downstream tasks \cite{b39,b40,b41}. Consequently, we propose to extract the semantic features corresponding to each binary mask, which could potentially enhance the accuracy of mask class association. Specifically, to ensure methodological consistency, we employ the image encoder component of the vanilla SAM for the extraction of semantic features from the image. Given an arbitrary resolution RS image $I_{i}$, we firstly rescale it into the $1024 \times 1024$ resolution to fit the input requirement of the image encoder~(ViT-H \cite{vit}), and the feature map $F_{i} \in \mathbb{R}^{256\times 64 \times 64}$ is extracted by the ViT-H. Then we design a feature alignment module to upsample the feature map by bilinear interpolation, thereby aligning it with the dimensions of the binary mask $M_{i} \in \mathbb{R}^{1\times 256\times 256}$. 

\noindent \textbf{Feature Clustering}: Based on the aligned feature map $\hat{F}_{i} \in \mathbb{R}^{256\times 256 \times 256}$, we propose a PFL mapping module, which uses the binary mask $M_{i} \in \mathbb{R}^{1\times 256\times 256}$ to execute the conjunction with the aligned feature map $\hat{F}_{i}$ to obtain the PFL $\mathcal{L}_{i}$ for each binary mask, defined as Eq. \ref{eq:equation_5}:

\begin{equation}
 \label{eq:equation_5}
    \mathcal{L}_{i} = \hat{F}_{i} \odot M_{i}
 \end{equation}

\noindent where $\odot$ represents the element-wise product. Based on the aforementioned process, we can obtain the PFL associated with each binary mask, i.e., construct a PFL set $\left\{ \mathcal{L}_{i} \right\}_{i=0}^{N}$. However, given the substantial volume of data, often in the order of hundreds of thousands or even millions of images, the direct application of conventional clustering algorithms \cite{b42, b43} may result in complications such as I/O crashes and memory overflow.
To resolve this issue, we adopt a learnable online K-means \cite{k-means} to split the images into mini-batches to iteratively cluster the PFLs into $\mathcal{N}$ clusters, where $\mathcal{N}$ is set for different datasets, e.g., iSAID: $\mathcal{N}=16$ , SAMRS: $\mathcal{N}=64$. After completing the feature clustering, we can use the trained clustering model to classify each PCL based on the clustering centers and thus obtain the class of each binary mask. This allows us to predict the classes of all binary masks in the binary mask set, ultimately obtaining the pseudo-label of the image.
Moreover, we also provide a scheme for predicting the number of clusters for large-scale datasets without prior knowledge. Specifically, DeepDPM \cite{deepdpm} can offer an elegant, data-adaptive, and mathematically-principled solution for clustering when $\mathcal{N}$ is unknown. Thus, it can be easily incorporated in our ALPS framework that rely on clustering.

Please be advised that while k-means clustering is a well-explored method, its specific application in conjunction with SAM for RS images and the novel mechanism for aligning pseudo-labels represents a noteworthy adaptation and refinement for RS contexts.


\section{Pseudo-Labeled RS Datasets}
This section presents the pseudo-labeled datasets generated by the proposed pipeline for RS pre-training.

\subsection{iSAID-PL}
\begin{figure*}[tb]
  \subfloat[Some examples of Pseudo-Labels on iSAID-PL dataset.]{
  \includegraphics[width=1\textwidth]{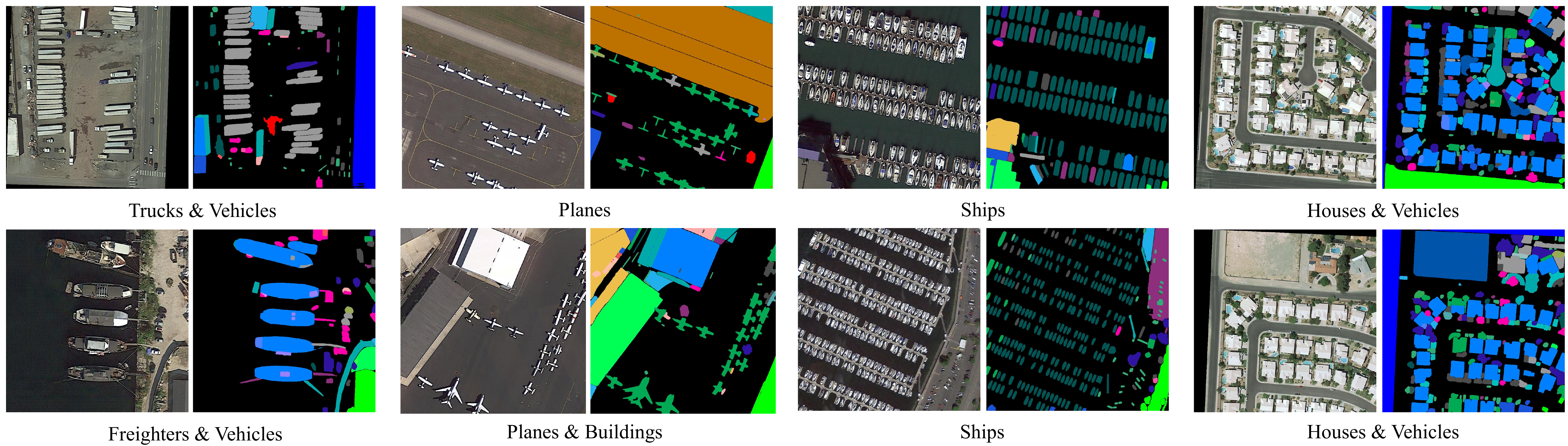}
  \label{fig:isaid}
  }
  \hfil
  \subfloat[Some examples of Pseudo-Labels on SAMRS-PL dataset.]{
  \includegraphics[width=1\textwidth]{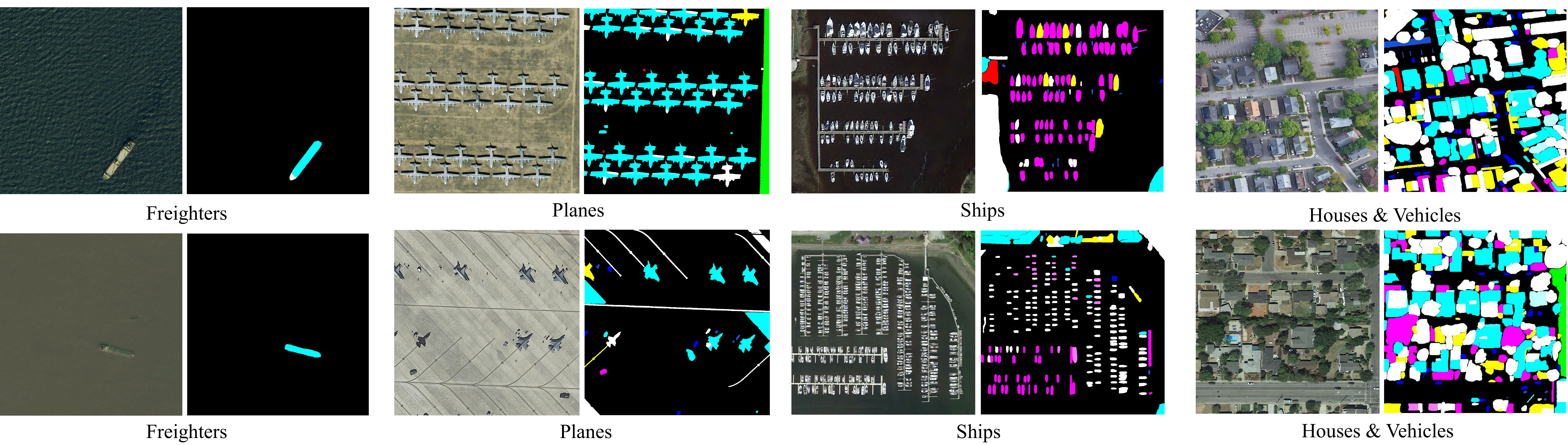}
  \label{fig:samrs}
  }
  \caption{Since we have used many types of colors from CoCo \cite{coco}, there may be very similar colors, but the classes they each represent are different. Moreover, these pseudo-labeled RS datasets adopt different random color sets.}
\end{figure*}

For the partition of the iSAID-PL dataset, we follow the original iSAID \cite{isaid}, i.e., 1411 training images, 458 validating images, and 937 testing images. 
The size of the images remains consistent with the original size. Based on the above original images, we use mmsegmentation \cite{mmseg} to crop the image, 
generating 33,978 training image patches, 11,644 validating image patches, and 937 testing image patches. 
For pseudo-labels, we adjust the pixels of the area that has not been covered to 255 by default. We set the number of classes $\mathcal{N}$ to 16 to align with iSAID, and the pixel color corresponding to each category is also one-to-one.
Moreover, in order to have a more intuitive understanding of our iSAID-PL, we have visualized some pseudo-labels. 
As shown in Fig \ref{fig:isaid}, we can discern that our methodology is capable of efficaciously annotating the majority of planes as green, ships as dark green, freighters as light blue, vehicles as pink, trucks as gray, and houses as dark blue. This superior visualization can be attributed to our feature clustering predicated on PFLs, which facilitates the allocation of identical classes across disparate RS images.


\subsection{SAMRS-PL}
Different from the partition of the SAMRS, we merge the Fast, SIOR, and SOTA datasets into a single comprehensive dataset for processing. This dataset comprises a total of 105,086 images. We employ a 7:3 ratio to partition the dataset, resulting in 73,560 images for training and 31,526 images for validation. Given that our methodology solely utilizes SAMRS-PL for pre-training, we do not allocate a separate test set. 
We utilize the ALPS to predict the pseudo-label for each image. 
For pseudo-label annotations, we adjust the pixels of the area that has not been covered to 255 by default. We set the classes of label $\mathcal{N}$ to 64 to align with SAMRS, and the pixel color corresponding to each category is also one-to-one.
As shown in Fig \ref{fig:samrs}, we have undertaken the visualization of pseudo-labels derived from the SAMRS-PL to foster a deeper comprehension of our methodology. We can observe that our method can effectively annotating the majority of freighters as indigo, ships as pink, planes as light blue and small vehicles as dark blue. Since we have set the $\mathcal{N}$ to 46, which causes the similar color masks within different classes, but these colors is similar but different.
The significant visualization can be attributed to our feature clustering predicated on PFLs, which streamlines the assignment of identical classes across different RS images. Nonetheless, it is observable that certain classes may exhibit multiple colors. This could potentially be attributed to an excessively large configuration of the parameter $\mathcal{N}$, and due to the inherent category differences between instances such as different ships, which subsequently engender disparities among high-level semantic features. However, as can be seen from the aforementioned figure, the annotations predominantly adhere to these colors, with no additional colors being introduced.

\begin{table*}[!t]
    \centering
    \resizebox{1\linewidth}{!}{
    \begin{tabular}{c|ccc|cccccccccccccccc|cc} 
    \toprule[1.2pt]
        Methods & Type & Pre-trained & Step & BG & SH & ST & BD & TC & BC & GTF & BR & LV & SV & HC & SP & RA & SBF & PL & HA & mIoU & mAcc \\
        \hline
        \multicolumn{2}{c}{\textbf{Convolution Based Network}} \\
        \hline
        \multirow{7}{*}{UperNet} 
            & Baseline & - & 80k & 98.67 & 68.80 & 71.29 & 71.98 & 86.67 & 52.68 & 57.35 & 39.94 & 63.29 & 48.30 & 0.00 & 48.09 & 47.35 & 72.38 & 82.74 & 58.20 & 64.52 & 73.70 \\
            & Overall & iSAID-PL & 80k & 98.61 & 68.28 & 72.05 & 71.74 & 86.79 & 52.44 & 56.58 & 40.25 & 63.26 & 48.56 & 0.00 & 54.67 & 47.29 & 70.50 & 82.93 & 59.58 & 64.90 & 73.96 \\
            & Backbone & iSAID-PL & 80k & 98.39 & 59.85 & 62.92 & 61.74 & 83.12 & 42.24 & 52.28 & 29.92 & 56.92 & 45.22 & 0.00 & 48.31 & 46.28 & 61.27 & 76.66 & 53.96 & 58.61 & 65.16 \\
            & Head & iSAID-PL & 80k & 98.67 & 70.20 & 70.34 & 76.33 & 88.18 & 61.91 & 56.54 & 40.37 & 63.67 & 49.37 & 0.00 & 50.92 & 58.53 & 73.83 & 83.68 & 57.56 & \textbf{66.67} & \textbf{74.38} \\
            & Overall & SAMRS-PL & 80k & 98.65 & 67.83 & 71.23 & 69.63 & 87.13 & 52.65 & 57.32 & 40.01 & 63.20 & 45.76 & 0.00 & 52.58 & 49.87 & 73.77 & 82.38 & 57.80 & 64.65 & 74.17 \\
            & Backbone & SAMRS-PL & 80k & 98.42 & 58.44 & 63.38 & 67.39 & 84.42 & 42.98 & 45.68 & 29.86 & 56.85 & 33.61 & 0.00 & 45.76 & 40.94 & 65.49 & 78.43 & 50.37 & 57.47 & 64.68 \\
            & Head & SAMRS-PL & 80k & 98.70 & 70.59 & 72.46 & 75.77 & 88.52 & 52.30 & 56.85 & 41.33 & 63.29 & 49.62 & 0.00 & 47.28 & 59.24 & 73.87 & 83.87 & 57.86 & 66.10 & 73.51 \\
        \hline
        \multirow{7}{*}{PspNet}  
            & Baseline & - & 80k & 98.66 & 69.03 & 72.52 & 73.27 & 87.93 & 64.52 & 56.83 & 37.98 & 62.41 & 45.01 & 0.00 & 50.46 & 55.62 & 76.01 & 82.10 & 54.30 & 61.67 & 74.41 \\
            & Overall & iSAID-PL & 80k & 98.70 & 69.84 & 73.16 & 73.93 & 88.68 & 65.50 & 57.18 & 41.53 & 63.08 & 46.91 & 0.00 & 51.93 & 62.49 & 76.78 & 83.30 & 56.43 & 63.09 & 71.56 \\
            & Backbone & iSAID-PL & 80k & 88.22 & 57.43 & 68.45 & 66.44 & 72.19 & 57.33 & 40.82 & 36.64 & 55.13 & 30.84 & 0.00 & 42.57 & 50.94 & 66.58 & 75.91 & 43.02 & 53.17 & 62.74 \\
            & Head & iSAID-PL & 80k & 98.64 & 69.32 & 72.85 & 72.13 & 87.46 & 62.73 & 56.63 & 40.97 & 62.48 & 46.56 & 0.00 & 51.40 & 62.00 & 75.89 & 82.61 & 56.61 & 62.39 & 71.99 \\
            & Overall & SAMRS-PL & 80k & 98.61 & 70.25 & 74.41 & 75.10 & 88.94 & 65.64 & 57.22 & 41.04 & 67.42 & 47.14 & 0.00 & 54.21 & 63.20 & 77.67 & 88.03 & 58.73 & \textbf{64.37} & \textbf{75.00} \\
            & Backbone & SAMRS-PL & 80k & 98.15 & 62.62 & 70.50 & 65.47 & 77.15 & 57.50 & 44.79 & 37.72 & 58.86 & 35.51 & 0.00 & 41.22 & 41.70 & 70.22 & 76.04 & 45.22 & 55.17 & 68.41 \\
            & Head & SAMRS-PL & 80k & 98.64 & 68.13 & 73.27 & 75.54 & 87.79 & 54.28 & 60.11 & 42.53 & 62.39 & 47.35 & 0.00 & 42.34 & 64.02 & 75.93 & 85.16 & 54.80 & 62.02 & 74.05 \\
        \hline
        \multirow{7}{*}{HRNet}  
            & Baseline & - & 80k & 98.71 & 68.80 & 67.55 & 74.43 & 87.82 & 58.66 & 55.16 & 42.36 & 64.03 & 50.24 & 0.00 & 48.57 & 62.93 & 77.74 & 84.30 & 59.37 & 62.54 & 73.82 \\
            & Overall & iSAID-PL & 80k & 98.66 & 69.13 & 67.65 & 74.03 & 87.53 & 61.68 & 59.78 & 44.96 & 62.60 & 49.11 & 0.00 & 47.39 & 61.11 & 73.25 & 83.02 & 60.15 & 62.50 & 72.18 \\
            & Backbone & iSAID-PL & 80k & 98.41 & 58.12 & 54.29 & 60.49 & 82.88 & 42.54 & 43.50 & 32.65 & 54.93 & 45.95 & 0.00 & 43.12 & 56.36 & 66.94 & 78.27 & 59.89 & 54.27 & 59.30 \\
            & Head & iSAID-PL & 80k & 98.71 & 69.41 & 72.14 & 75.69 & 87.67 & 61.36 & 58.07 & 41.33 & 62.91 & 49.58 & 0.00 & 47.81 & 69.01 & 78.20 & 83.92 & 59.73 & 63.47 & 74.50 \\
            & Overall & SAMRS-PL & 80k & 98.66 & 69.13 & 67.65 & 74.03 & 87.53 & 61.68 & 59.78 & 44.96 & 62.60 & 49.11 & 0.00 & 47.39 & 65.11 & 73.25 & 83.02 & 60.15 & 62.75 & 73.32 \\
            & Backbone & SAMRS-PL & 80k & 98.41 & 58.39 & 56.16 & 60.00 & 82.66 & 46.82 & 43.44 & 32.73 & 54.58 & 45.52 & 0.00 & 47.07 & 60.12 & 60.16 & 78.00 & 60.28 & 55.27 & 62.10 \\
            & Head & SAMRS-PL & 80k & 98.70 & 67.77 & 71.98 & 74.21 & 87.87 & 63.71 & 59.03 & 42.51 & 63.62 & 51.40 & 0.00 & 50.44 & 67.89 & 75.21 & 84.57 & 58.58 & \textbf{63.59} & \textbf{74.64} \\
        \hline
        \multirow{7}{*}{Deeplabv3plus}  
            & Baseline & - & 80k & 98.62 & 70.07 & 69.55 & 58.06 & 88.33 & 60.25 & 56.44 & 41.11 & 64.11 & 47.49 & 0.00 & 51.70 & 42.52 & 72.34 & 81.27 & 58.34 & 60.01 & 74.27 \\
            & Overall & iSAID-PL & 80k & 98.62 & 66.61 & 71.80 & 66.93 & 87.06 & 59.82 & 57.65 & 36.52 & 62.61 & 46.79 & 0.00 & 57.22 & 47.14 & 70.89 & 80.18 & 57.68 & 60.46 & 73.94 \\
            & Backbone & iSAID-PL & 80k & 98.23 & 54.05 & 54.80 & 53.80 & 78.72 & 46.86 & 55.82 & 28.27 & 52.70 & 42.02 & 0.00 & 49.30 & 36.60 & 55.04 & 74.81 & 47.77 & 51.79 & 59.39 \\
            & Head & iSAID-PL & 80k & 98.67 & 70.72 & 73.13 & 70.72 & 87.82 & 59.99 & 52.30 & 42.13 & 63.30 & 48.97 & 0.00 & 52.33 & 47.30 & 75.61 & 82.86 & 58.62 & 61.52 & 74.84 \\
            & Overall & SAMRS-PL & 80k & 98.67 & 69.94 & 72.04 & 70.19 & 86.39 & 57.92 & 60.17 & 38.48 & 64.46 & 46.17 & 0.00 & 47.22 & 45.44 & 73.39 & 83.53 & 55.59 & 60.60 & 73.92 \\
            & Backbone & SAMRS-PL & 80k & 98.24 & 54.41 & 55.69 & 60.99 & 78.17 & 48.71 & 59.53 & 32.82 & 53.87 & 43.58 & 0.00 & 51.86 & 33.84 & 51.73 & 75.96 & 47.58 & 52.93 & 60.87 \\
            & Head & SAMRS-PL & 80k & 98.72 & 70.88 & 72.50 & 75.65 & 88.44 & 66.60 & 59.49 & 42.18 & 64.98 & 50.44 & 0.00 & 50.74 & 62.11 & 77.04 & 84.01 & 55.82 & \textbf{63.72} & \textbf{75.13} \\
            \hline
        \multicolumn{2}{c}{\textbf{Transformer Based Network}} \\
            \hline
        \multirow{7}{*}{ViT}  
            & Baseline & - & 160k & 98.62 & 66.81 & 68.87 & 77.13 & 85.33 & 56.27 & 63.17 & 46.87 & 59.27 & 38.91 & 0.00 & 47.24 & 67.58 & 73.80 & 80.05 & 52.79 & 61.42 & 72.51 \\
            & Overall & iSAID-PL & 160k & 98.64 & 66.31 & 68.94 & 75.68 & 85.75 & 57.55 & 63.80 & 43.80 & 60.48 & 39.52 & 0.00 & 46.55 & 67.63 & 75.83 & 80.46 & 52.74 & 61.48 & 72.57 \\
            & Backbone & iSAID-PL & 160k & 98.63 & 66.17 & 69.19 & 75.43 & 85.51 & 57.53 & 60.88 & 38.59 & 60.42 & 39.76 & 0.00 & 48.24 & 67.80 & 74.01 & 80.19 & 52.66 & 60.94 & 72.43 \\
            & Head & iSAID-PL & 160k & 98.64 & 67.15 & 68.61 & 78.08 & 85.53 & 55.67 & 61.87 & 45.93 & 60.42 & 39.58 & 0.00 & 51.04 & 70.76 & 74.19 & 80.26 & 55.27 & \textbf{62.06} & \textbf{72.90} \\
            & Overall & SAMRS-PL & 160k & 98.63 & 66.66 & 69.11 & 76.44 & 84.49 & 60.41 & 59.47 & 45.26 & 60.01 & 38.48 & 0.00 & 49.97 & 67.11 & 73.84 & 80.38 & 55.71 & 61.62 & 72.41 \\
            & Backbone & SAMRS-PL & 160k & 98.63 & 66.74 & 69.11 & 77.42 & 85.51 & 57.70 & 60.63 & 42.68 & 60.34 & 38.78 & 0.00 & 47.94 & 65.88 & 74.36 & 80.14 & 53.64 & 61.22 & 72.16 \\
            & Head & SAMRS-PL & 160k & 98.62 & 66.14 & 68.75 & 78.30 & 85.86 & 56.79 & 61.52 & 46.89 & 58.68 & 39.63 & 0.00 & 47.44 & 68.31 & 73.97 & 79.89 & 54.04 & 61.55 & 72.83 \\
            \hline
        \multirow{7}{*}{Swin}  
            & Baseline & - & 160k & 98.74 & 71.25 & 73.93 & 80.11 & 88.61 & 61.76 & 60.36 & 39.38 & 64.27 & 48.93 & 0.00 & 49.32 & 66.71 & 77.30 & 84.32 & 59.03 & 64.00 & 75.30 \\
            & Overall & iSAID-PL & 160k & 98.75 & 71.17 & 73.62 & 81.76 & 88.76 & 63.00 & 60.29 & 39.48 & 64.71 & 49.39 & 0.00 & 49.52 & 66.59 & 76.34 & 84.43 & 60.23 & 64.25 & 75.10 \\
            & Backbone & iSAID-PL & 160k & 98.74 & 73.40 & 72.38 & 78.91 & 88.35 & 62.62 & 59.93 & 44.49 & 64.57 & 48.97 & 0.00 & 50.56 & 68.59 & 76.41 & 84.30 & 58.99 & 64.45 & 75.20 \\
            & Head & iSAID-PL & 160k & 98.75 & 72.08 & 74.14 & 78.32 & 88.56 & 64.15 & 60.49 & 46.18 & 64.65 & 49.51 & 0.00 & 51.14 & 67.72 & 77.01 & 84.43 & 58.88 & 64.75 & 75.40 \\
            & Overall & SAMRS-PL & 160k & 98.76 & 73.16 & 72.77 & 80.39 & 88.64 & 63.83 & 64.44 & 41.58 & 64.81 & 48.82 & 0.00 & 51.27 & 66.32 & 77.08 & 84.34 & 60.97 & \textbf{64.82} & \textbf{76.12} \\
            & Backbone & SAMRS-PL & 160k & 98.75 & 71.36 & 72.29 & 79.26 & 88.31 & 64.56 & 62.64 & 43.77 & 65.10 & 48.82 & 0.00 & 51.54 & 69.04 & 76.57 & 84.48 & 60.15 & 64.79 & 75.75 \\
            & Head & SAMRS-PL & 160k & 98.76 & 72.79 & 72.99 & 79.93 & 89.27 & 62.51 & 62.06 & 44.15 & 64.70 & 49.20 & 0.00 & 50.49 & 67.18 & 77.66 & 84.51 & 58.13 & 64.75 & 75.59 \\
            \hline
        \multirow{7}{*}{Segmenter}  
            & Baseline & - & 160k & 98.30 & 56.36 & 58.47 & 72.03 & 79.94 & 56.20 & 62.86 & 35.64 & 48.59 & 23.13 & 0.00 & 35.45 & 60.65 & 76.24 & 65.70 & 39.38 & 57.93 & 66.20 \\
            & Overall & iSAID-PL & 160k & 98.32 & 56.11 & 59.44 & 74.14 & 80.98 & 54.04 & 62.71 & 33.56 & 49.43 & 23.10 & 0.00 & 35.47 & 59.58 & 76.53 & 66.74 & 39.12 & 57.95 & 66.17 \\
            & Backbone & iSAID-PL & 160k & 98.31 & 55.23 & 59.19 & 75.53 & 81.11 & 54.41 & 61.75 & 32.43 & 50.09 & 22.86 & 0.00 & 27.51 & 60.92 & 74.60 & 66.77 & 38.69 & 57.29 & 65.92 \\
            & Head & iSAID-PL & 160k & 98.30 & 56.16 & 59.01 & 73.30 & 79.83 & 55.86 & 61.86 & 36.11 & 48.76 & 22.39 & 0.00 & 34.20 & 61.41 & 76.68 & 65.79 & 40.23 & 57.99 & \textbf{66.55} \\
            & Overall & SAMRS-PL & 160k & 98.32 & 55.98 & 58.56 & 74.65 & 80.19 & 54.80 & 61.79 & 33.31 & 50.24 & 23.32 & 0.00 & 32.27 & 65.85 & 75.51 & 66.38 & 41.44 & \textbf{58.17} & 66.37 \\
            & Backbone & SAMRS-PL & 160k & 98.29 & 55.38 & 58.03 & 75.15 & 81.22 & 55.81 & 61.69 & 33.10 & 50.79 & 22.89 & 0.00 & 28.02 & 61.33 & 74.17 & 66.43 & 38.32 & 57.37 & 66.28\\
            & Head & SAMRS-PL & 160k & 98.30 & 55.64 & 58.59 & 74.50 & 79.98 & 57.09 & 62.83 & 38.09 & 49.15 & 21.60 & 0.00 & 33.81 & 60.27 & 75.69 & 65.63 & 40.37 & 58.10 & 66.36 \\
            \hline
        \multirow{7}{*}{Segformer}  
            & Baseline & - & 160k & 98.73 & 70.99 & 74.07 & 79.18 & 88.53 & 64.12 & 66.01 & 43.82 & 64.23 & 46.15 & 0.00 & 47.12 & 71.58 & 77.48 & 83.55 & 58.09 & 64.60 & 75.62 \\
            & Overall & iSAID-PL & 160k & 98.75 & 72.05 & 73.49 & 79.33 & 88.30 & 66.92 & 65.58 & 41.90 & 64.41 & 46.38 & 0.00 & 46.79 & 75.34 & 77.52 & 83.52 & 57.99 & 64.89 & 75.97 \\
            & Backbone & iSAID-PL & 160k & 98.73 & 70.18 & 73.17 & 77.31 & 88.04 & 61.73 & 64.38 & 46.31 & 63.12 & 46.88 & 0.00 & 50.12 & 70.33 & 78.48 & 83.27 & 56.81 & 64.30 & 75.34 \\
            & Head & iSAID-PL & 160k & 98.74 & 70.57 & 73.39 & 79.24 & 88.50 & 62.05 & 66.68 & 44.87 & 63.57 & 47.02 & 0.00 & 49.76 & 73.56 & 78.30 & 83.70 & 57.77 & 64.86 & 75.86 \\
            & Overall & SAMRS-PL & 160k & 98.76 & 72.32 & 73.57 & 80.03 & 89.05 & 64.76 & 65.6 & 46.68 & 63.05 & 46.75 & 0.00 & 46.98 & 75.53 & 78.91 & 83.67 & 59.84 & \textbf{65.35} & \textbf{76.53} \\
            & Backbone & SAMRS-PL & 160k & 98.76 & 72.67 & 72.81 & 80.57 & 88.14 & 63.45 & 65.50 & 41.55 & 63.68 & 46.55 & 0.00 & 48.53 & 72.16 & 78.78 & 83.70 & 58.33 & 64.70 & 75.97 \\
            & Head & SAMRS-PL & 160k & 98.74 & 71.25 & 73.82 & 79.47 & 88.32 & 64.87 & 62.61 & 45.26 & 63.31 & 46.09 & 0.00 & 46.82 & 73.43 & 79.61 & 83.63 & 59.96 & 64.82 & 75.60 \\
            \bottomrule
            \end{tabular}
    }
    \caption{Segmentation results of different methods on the iSAID dataset \cite{isaid}. BG: background. SH: ship. ST: storage tank. BD: baseball diamond. TC: tennis court. BC: baseball court. GTF: ground track field. BR: bridge. LV: large vehicle. SV: small vehicle. HC: helicopter. SP: swimming pool. RA: roundabout. SBF: soccer ball field. PL: plane. HA: harbor.}
    \label{tab:table.2}
\end{table*}

\section{Experiment}
To evaluate the effectiveness of our generated pseudo-labeled RS datasets, we have utilized these datasets for pre-training. The performance of the pre-trained models are then validated across various datasets, e.g., iSAID \cite{isaid}, ISPRS Potsdam \cite{potsdam}. 

\subsection{Experimental Settings}
\label{5.1}
We use the mmsegmentaion \cite{mmseg} framework and currently, eight mainstream methods are used as the baseline for pre-training and fine-tuning, e.g., UperNet \cite{upernet}, PspNet \cite{pspnet}, HRNet \cite{hrnet}, Deeplabv3plus \cite{deeplabv3plus}, ViT \cite{vit}, Swin-Transformer \cite{swin}, SegMenter \cite{segmenter}, SegFormer \cite{segformer}. The initial phase of pre-training is executed on the iSAID-PL and SAMRS-PL datasets respectively. Subsequently, fine-tuning is performed on the iSAID and ISPRS Potsdam datasets. This comprehensive process serves to verify the effectiveness of our constructed datasets. Specifically, the specific types of models we use are as follows: UperNet-r50, PspNet-r50, FCN-HRNet-r48, Deeplabv3plus-r50b, ViT-B, Swin-Base, Segmenter-ViT-B, Segformer-MiT-B5. Moreover, in the RS community, iSAID and ISPRS Potsdam have been widely used annotated datasets for evaluating segmentation methods, and we use them to assess the pre-trained models. All experiments are conducted using the PyTorch framework on a setup with 2x NVIDIA GeForce RTX 4090 GPUs under fair configurations. 


\subsection{Benchmarking on iSAID Dataset}
In this section, we have compared the aforementioned methods on iSAID dataset to validate the effectiveness of our pre-training scheme, where are pre-trained on our proposed pseudo-labeled RS datasets. To provide more comprehensive quantitative results, we have divided our pre-trained model into three pre-trained sub-models: overall, backbone, and head. During the fine-tuning phase, we loaded the aforementioned sub-models as initial parameters and then trained them under the same environment and configuration to compare the results with the corresponding baselines. All of the baselines are fine-tuned with the official pre-trained backbones loaded by mmsegmentaion. The quantitative results are shown in Table \ref{tab:table.2}, we can observe that following pre-training on pseudo-labeled RS datasets, the fine-tuning results on iSAID surpass the baseline results, e.g., iSAID-PL \& Overall: UperNet: 64.90 vs. 64.52, Segformer: 64.89 vs. 64.60, underscoring the efficacy of our predicted pseudo-labels in positively improving the initialization of parameters. We attribute the significant improvement to the integration of segmentation pre-training with pseudo-labels, which provides a valuable prior for segmentation. However, the sub-model of the backbone exhibits relatively weaker performance compared to others. We believe this is due to insufficient pre-training steps. Therefore, we have further explored using a larger number of pre-training steps to validate our analysis, as discussed in Sec \ref{sec:5.4}.

\begin{table*}[!t]
    \centering
    \scriptsize
    \subfloat{
    \resizebox{0.5\linewidth}{!}{
        \begin{tabular}{c|ccc|cccccc|cc} 
        \toprule[1.2pt]
            Methods & Type & Pre-trained & Step & IS & BD & LV & TR & CR & CL & mIoU & mAcc \\
            \hline
            \multicolumn{3}{c}{\textbf{Convolution Based Network}} \\
            \hline
            \multirow{7}{*}{UperNet} 
                & Baseline & - & 80k & 81.06 & 89.02 & 70.57 & 74.39 & 74.17 & 37.88 & 71.18 & 83.74 \\
                & Overall & iSAID-PL & 80k & 80.73 & 89.84 & 69.77 & 73.99 & 74.67 & 38.52 & 71.25 & 85.07 \\
                & Backbone & iSAID-PL & 80k & 80.62 & 89.70 & 70.04 & 73.78 & 73.26 & 37.64 & 70.84 & 84.78 \\
                & Head & iSAID-PL & 80k & 81.17 & 89.98 & 70.34 & 74.46 & 74.18 & 38.19 & 71.39 & 85.33 \\
                & Overall & SAMRS-PL & 80k & 81.08 & 90.10 & 70.15 & 74.47 & 74.38 & 37.55 & 71.29 & 85.10 \\
                & Backbone & SAMRS-PL & 80k & 80.86 & 89.58 & 69.78 & 74.12 & 74.10 & 37.77 & 71.03 & 84.95 \\
                & Head & SAMRS-PL & 80k & 80.90 & 90.91 & 69.10 & 73.32 & 73.89 & 38.71 & \textbf{71.45} & \textbf{87.12} \\
            \hline
            \multirow{7}{*}{Pspnet} 
                & Baseline & - & 80k & 81.13 & 89.80 & 70.30 & 74.44 & 73.88 & 37.39 & 71.16 & 85.15 \\
                & Overall & iSAID-PL & 80k & 80.98 & 89.78 & 70.62 & 74.48 & 73.72 & 39.24 & 71.47 & 85.39 \\
                & Backbone & iSAID-PL & 80k & 78.02 & 88.33 & 69.98 & 73.75 & 72.90 & 36.41 & 69.89 & 80.38 \\
                & Head & iSAID-PL & 80k & 81.15 & 90.13 & 70.20 & 74.35 & 75.10 & 38.94 & \textbf{71.64} & \textbf{85.46} \\
                & Overall & SAMRS-PL & 80k & 80.94 & 89.65 & 70.50 & 74.47 & 73.95 & 38.89 & 71.40 & 85.37 \\
                & Backbone & SAMRS-PL & 80k & 79.99 & 88.61 & 70.02 & 73.60 & 73.83 & 37.30 & 70.56 & 84.57\\
                & Head & SAMRS-PL & 80k & 81.13 & 90.02 & 70.07 & 74.35 & 74.41 & 38.76 & 71.46 & 85.40 \\
            \hline
            \multirow{7}{*}{HRNet} 
                & Baseline & - & 80k & 80.59 & 89.44 & 69.74 & 73.63 & 72.96 & 37.41 & 70.63 & 84.86 \\
                & Overall & iSAID-PL & 80k & 80.27 & 89.33 & 70.09 & 73.92 & 73.87 & 39.54 & 71.17 & 85.25 \\
                & Backbone & iSAID-PL & 80k & 80.37 & 88.99 & 69.47 & 73.35 & 73.30 & 38.11 & 70.60 & 84.76 \\
                & Head & iSAID-PL & 80k & 80.95 & 89.96 & 70.70 & 74.74 & 73.93 & 37.62 & 71.30 & 85.12 \\
                & Overall & SAMRS-PL & 80k & 80.83 & 89.98 & 70.23 & 74.63 & 73.94 & 36.19 & 70.97 & \textbf{85.31} \\
                & Backbone & SAMRS-PL & 80k & 79.43 & 87.60 & 68.59 & 72.05 & 71.83 & 37.52 & 69.50 & 84.21 \\
                & Head & SAMRS-PL & 80k & 81.05 & 89.97 & 70.60 & 74.74 & 74.29 & 38.59 & \textbf{71.54} & 85.27 \\
            \hline
            \multirow{7}{*}{Deeplabv3plus} 
                & Baseline & - & 80k & 80.93 & 90.17 & 69.91 & 74.32 & 74.77 & 37.45 & 71.26 & 85.02 \\
                & Overall & iSAID-PL & 80k & 81.07 & 90.06 & 70.33 & 74.25 & 73.71 & 39.44 & \textbf{71.48} & \textbf{85.53} \\
                & Backbone & iSAID-PL & 80k & 81.26 & 93.35 & 70.06 & 74.67 & 74.26 & 38.22 & 71.39 & 85.30 \\
                & Head & iSAID-PL & 80k & 80.86 & 90.03 & 70.52 & 74.32 & 73.46 & 39.37 & 71.43 & 85.31 \\
                & Overall & SAMRS-PL & 80k & 80.97 & 90.07 & 69.71 & 73.95 & 73.76 & 39.33 & 71.29 & 85.05 \\
                & Backbone & SAMRS-PL & 80k & 80.42 & 88.86 & 68.89 & 73.43 & 72.51 & 36.90 & 70.17 & 84.41 \\
                & Head & SAMRS-PL & 80k & 81.30 & 90.37 & 69.93 & 74.46 & 73.82 & 38.02 & 71.35 & 85.16 \\
            \bottomrule
         \end{tabular}
    }
    }
    \subfloat{
    \resizebox{0.49\linewidth}{!}{
        \begin{tabular}{c|ccc|cccccc|cc} 
        \toprule[1.2pt]
            Methods & Type & Pre-trained & Step & IS & BD & LV & TR & CR & CL & mIoU & mAcc \\
            \hline
            \multicolumn{3}{c}{\textbf{Transformer Based Network}} \\
            \hline
            \multirow{7}{*}{ViT} 
                & Baseline & - & 160k & 80.10 & 89.42 & 70.39 & 73.65 & 73.49 & 35.52 & 70.25 & 84.26 \\
                & Overall & iSAID-PL & 160k & 80.51 & 89.53 & 70.48 & 73.50 & 73.05 & 35.88 & 70.49 & 84.46 \\
                & Backbone & iSAID-PL & 160k & 80.46 & 89.20 & 70.39 & 73.59 & 73.09 & 35.29 & 70.34 & 84.32 \\
                & Head & iSAID-PL & 160k & 81.14 & 89.95 & 70.30 & 73.56 & 73.13 & 37.05 & \textbf{70.85} & \textbf{84.77} \\
                & Overall & SAMRS-PL & 160k & 80.83 & 89.66 & 69.98 & 73.22 & 73.24 & 34.88 & 70.30 & 84.24 \\
                & Backbone & SAMRS-PL & 160k & 80.54 & 89.39 & 70.28 & 73.55 & 73.25 & 35.21 & 70.37 & 84.34 \\
                & Head & SAMRS-PL & 160k & 81.34 & 89.73 & 70.29 & 73.55 & 72.85 & 36.96 & 70.79 & \textbf{84.77} \\
            \hline
            \multirow{7}{*}{Swin} 
                & Baseline & - & 160k & 81.21 & 90.03 & 71.04 & 74.72 & 73.64 & 37.11 & 71.29 & 85.11 \\
                & Overall & iSAID-PL & 160k & 81.71 & 90.22 & 71.15 & 74.77 & 73.94 & 39.78 & 71.93 & 85.61 \\
                & Backbone & iSAID-PL & 160k & 81.15 & 90.01 & 70.96 & 74.70 & 73.65 & 38.21 & 71.45 & 85.33 \\
                & Head & iSAID-PL & 160k & 81.50 & 90.30 & 70.93 & 74.92 & 73.90 & 37.82 & 71.56 & 85.30 \\
                & Overall & SAMRS-PL & 160k & 81.34 & 90.23 & 70.86 & 74.73 & 73.55 & 37.7 & 71.40 & 85.21 \\
                & Backbone & SAMRS-PL & 160k & 81.55 & 90.31 & 70.92 & 74.68 & 73.93 & 38.91 & 71.72 & 85.49 \\
                & Head & SAMRS-PL & 160k & 81.58 & 90.27 & 71.39 & 74.98 & 74.27 & 39.52 & \textbf{72.00} & \textbf{85.68} \\
            \hline
            \multirow{7}{*}{Segmenter} 
                & Baseline & - & 160k & 80.94 & 89.14 & 70.95 & 74.28 & 71.95 & 37.18 & 70.74 & 84.76 \\
                & Overall & iSAID-PL & 160k & 80.47 & 89.99 & 71.05 & 74.32 & 72.39 & 37.15 & \textbf{70.90} & 84.60 \\
                & Backbone & iSAID-PL & 160k & 80.72 & 89.30 & 70.73 & 74.29 & 72.12 & 36.70 & 70.64 & 84.60 \\
                & Head & iSAID-PL & 160k & 80.62 & 89.77 & 70.95 & 74.26 & 72.33 & 36.49 & 70.74 & 84.54 \\
                & Overall & SAMRS-PL & 160k & 81.02 & 89.27 & 70.72 & 74.00 & 71.97 & 37.68 & 70.78 & 84.71 \\
                & Backbone & SAMRS-PL & 160k & 80.63 & 89.16 & 70.80 & 74.08 & 71.80 & 37.62 & 70.68 & 84.72 \\
                & Head & SAMRS-PL & 160k & 80.83 & 89.57 & 71.13 & 74.21 & 71.91 & 37.74 & \textbf{70.90} & \textbf{84.83} \\
            \hline
            \multirow{7}{*}{Segformer} 
                & Baseline & - & 160k & 81.58 & 90.09 & 71.41 & 74.98 & 73.46 & 37.48 & 71.50 & 84.16 \\
                & Overall & iSAID-PL & 160k & 81.78 & 90.48 & 71.47 & 75.11 & 73.83 & 39.93 & \textbf{72.10} & \textbf{85.72} \\
                & Backbone & iSAID-PL & 160k & 81.58 & 89.96 & 71.36 & 74.96 & 73.53 & 36.80 & 71.37 & 84.89\\
                & Head & iSAID-PL & 160k & 81.89 & 90.52 & 71.62 & 75.24 & 73.81 & 38.52 & 71.93 & 84.52 \\
                & Overall & SAMRS-PL & 160k & 81.81 & 90.45 & 71.13 & 74.87 & 73.75 & 38.86 & 71.81 & 85.49 \\
                & Backbone & SAMRS-PL & 160k & 81.82 & 90.36 & 71.13 & 74.88 & 73.51 & 38.77 & 71.74 & 85.52 \\
                & Head & SAMRS-PL & 160k & 82.06 & 90.60 & 71.55 & 75.21 & 73.61 & 41.12 & 71.86 & 85.56 \\
            \bottomrule
         \end{tabular}
    }
    }
    \caption{Segmentation results of different methods on the ISPRS Potsdam dataset \cite{potsdam}. IS: impervious surface. BD: building. LV: low vegetation. TR:tree. CR: car. CL: clutter}
    \label{tab:table.3}
\end{table*}



\subsection{Benchmarking on ISPRS Potsdam Dataset}
Furthermore, to validate the generalization of our constructed pseudo-labeled RS datasets, we have adopted the ISPRS Potsdam dataset during the fine-tuning phrase to mitigate potential issues such as overfitting or diminished generalizability that may arise due to the utilization of images from the identical domain in both the iSAID and iSAID-PL. The quantitative results are presented in Table \ref{tab:table.3}. Contrary to the fine-tuning results on iSAID, the performance of each method on the ISPRS Potsdam dataset is more closely aligned. And we can observe that all of methods demonstrate enhanced performance relative to their respective baselines, e.g., SAMRS-PL \& Overall: UperNet: 71.29 vs. 71.18, Segformer: 71.81 vs. 71.50. Moreover, it is noteworthy that in the ISPRS Potsdam dataset, the performance disparity between the backbone sub-model and other sub-models is not substantial like results on iSAID. 
Based on the evaluation results on iSAID and Potsdam datasets, we can conduct the effectiveness and generalization of our proposed pseudo-labeled RS datasets in downstream tasks, which means the annotations with precise labels will yield substantial benefits in pre-training, thereby enhancing the performance of fine-tuning. Besides, in contrast to the pesudo-label annotations, we also design a binary mask case to verify the label-level annotations is better than binary annotations, which will be analyzed in Sec \ref{sec:5.4}.




\begin{figure*}[t]
    \centering
    \subfloat{
        \centering
        \includegraphics[width=0.5\textwidth]{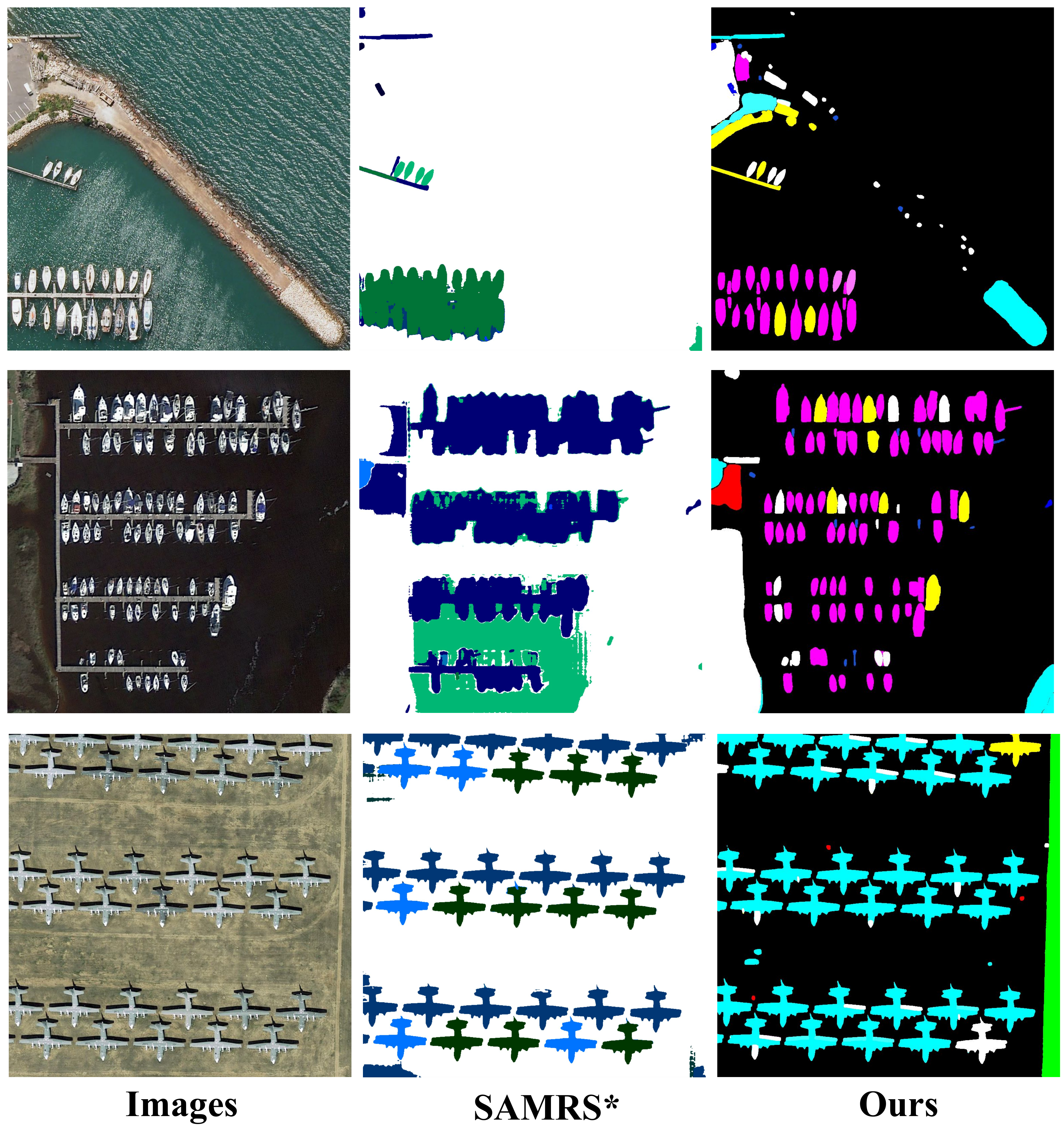}
        \label{fig:samrs_comp}
    }
    \subfloat{
      \centering
      \includegraphics[width=0.5\textwidth]{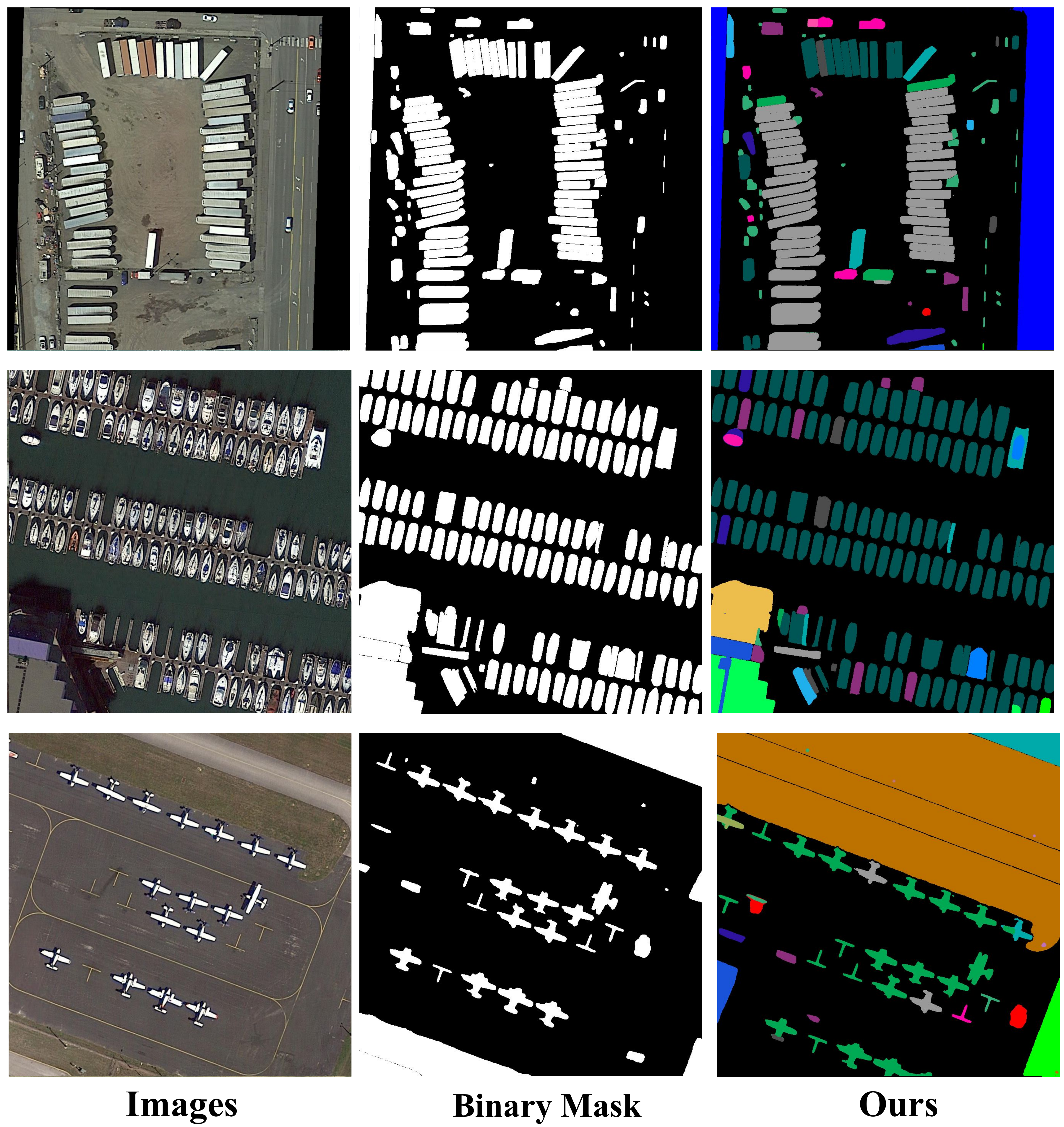}
      \label{fig:binary}
    }
    \caption{(a)~Qualitative results of different pre-training datasets constrcuted by different unsupervised manners. (b)~Some examples of binary masks generated by vanilla SAM. }
\end{figure*}

\begin{table*}[!t]
    \centering
    \scriptsize
    \subfloat{
    \resizebox{1\linewidth}{!}{
        \begin{tabular}{c|ccc|cccccccccccccccc|cc}  
        \toprule[1.2pt]
        Methods & Type & Pre-trained & Step & BG & SH & ST & BD & TC & BC & GTF & BR & LV & SV & HC & SP & RA & SBF & PL & HA & mIoU & mAcc \\
        \hline
        \multirow{4}{*}{UperNet} 
            & Backbone & SAMRS* & 80k & 98.22 & 57.99 & 58.38 & 50.35 & 82.21 & 41.54 & 45.67 & 26.33 & 52.91 & 35.23 & 0.00 & 31.75 & 38.87 & 51.51 & 76.56 & 51.95 & 53.30 & 63.50 \\
            & Backbone & SAMRS-PL & 80k & 98.42 & 58.44 & 63.38 & 67.39 & 84.42 & 42.98 & 45.68 & 29.86 & 56.85 & 33.61 & 0.00 & 45.76 & 40.94 & 65.49 & 78.43 & 50.37 & \textbf{57.47} & \textbf{64.68} \\
            \cline{2-22}
            & Overall & SAMRS* & 80k & 98.35 & 58.91 & 63.59 & 54.82 & 81.15 & 46.19 & 49.68 & 32.04 & 59.83 & 41.51 & 0.00 & 52.23 & 42.45 & 69.14 & 75.62 & 56.17 & 58.78 & 67.67 \\
            & Overall & SAMRS-PL & 80k & 98.65 & 67.83 & 71.23 & 69.63 & 87.13 & 52.65 & 57.32 & 40.01 & 63.20 & 45.76 & 0.00 & 52.58 & 49.87 & 73.77 & 82.38 & 57.80 & \textbf{64.65} & \textbf{74.17} \\
        \hline
        \multirow{4}{*}{Segformer} 
            & Backbone & SAMRS* & 160k & 98.73 & 70.31 & 73.40 & 79.52 & 88.01 & 63.13 & 63.73 & 47.40 & 63.47 & 46.03 & 0.00 & 47.82 & 65.59 & 76.93 & 83.53 & 59.40 & 64.19 & 75.04 \\
            & Backbone & SAMRS-PL & 160k & 98.76 & 72.67 & 72.81 & 80.57 & 88.14 & 63.45 & 65.50 & 41.55 & 63.68 & 46.55 & 0.00 & 48.53 & 72.16 & 78.78 & 83.70 & 58.33 & \textbf{64.70} & \textbf{75.97} \\
            \cline{2-22}
            & Overall & SAMRS* & 160k & 98.73 & 70.79 & 72.95 & 79.91 & 88.00 & 64.28 & 66.32 & 48.11 & 63.83 & 46.94 & 0.00 & 50.10 & 66.12 & 77.93 & 83.62 & 59.07 & 64.80 & 75.75 \\
            & Overall & SAMRS-PL & 160k & 98.76 & 72.32 & 73.57 & 80.03 & 89.05 & 64.76 & 65.6 & 46.68 & 63.05 & 46.75 & 0.00 & 46.98 & 75.53 & 78.91 & 83.67 & 59.84 & \textbf{65.35} & \textbf{76.53} \\    
            \bottomrule
         \end{tabular}
    }
    \label{tab:Table.6}
    }
    
    \subfloat{
    \resizebox{1\linewidth}{!}{
        \begin{tabular}{c|ccc|cccccccccccccccc|cc} 
        \toprule[1.2pt]
        Methods & Type & Pre-trained & Step & BG & SH & ST & BD & TC & BC & GTF & BR & LV & SV & HC & SP & RA & SBF & PL & HA & mIoU & mAcc \\
        \hline
        \multirow{4}{*}{UperNet} 
            & Backbone & Binary Mask & 80k & 96.98 & 25.20 & 0.00 & 0.00 & 0.00 & 0.00 & 0.00 & 0.00 & 19.82 & 0.12 & 0.00 & 0.00 & 0.00 & 0.00 & 0.23 & 17.81 & 10.03 & 11.97 \\
            & Backbone & iSAID-PL & 80k & 98.39 & 59.85 & 62.92 & 61.74 & 83.12 & 42.24 & 52.28 & 29.92 & 56.92 & 45.22 & 0.00 & 48.31 & 46.28 & 61.27 & 76.66 & 53.96 & 58.61 & 65.16 \\
            & Backbone & SAMRS-PL & 80k & 98.42 & 58.44 & 63.38 & 67.39 & 84.42 & 42.98 & 45.68 & 29.86 & 56.85 & 33.61 & 0.00 & 45.76 & 40.94 & 65.49 & 78.43 & 50.37 & 57.47 & 64.68 \\
            & Backbone & Mixed-PL  & 80k & 98.08 & 59.33 & 67.74 & 67.85 & 87.00 & 53.64 & 60.64 & 52.13 & 59.82 & 47.69 & 0.00 & 50.05 & 49.14 & 68.48 & 78.86 & 54.39 & \textbf{59.68} & \textbf{68.33}  \\
        \hline
        \multirow{4}{*}{Segformer} 
            & Backbone & Binary Mask & 160k & 97.10 & 30.30 & 0.88 & 0.00 & 0.00 & 0.00 & 0.00 & 0.00 & 21.98 & 3.21 & 0.00 & 0.00 & 0.00 & 0.00 & 2.79 & 19.88 & 11.01 & 13.55 \\
            & Backbone & iSAID-PL & 160k & 98.73 & 70.18 & 73.17 & 77.31 & 88.04 & 61.73 & 64.38 & 46.31 & 63.12 & 46.88 & 0.00 & 50.12 & 70.33 & 78.48 & 83.27 & 56.81 & 64.30 & 75.34 \\
            & Backbone & SAMRS-PL & 160k & 98.76 & 72.67 & 72.81 & 80.57 & 88.14 & 63.45 & 65.50 & 41.55 & 63.68 & 46.55 & 0.00 & 48.53 & 72.16 & 78.78 & 83.70 & 58.33 & 64.70 & 75.97 \\
            & Backbone & Mixed-PL  & 160k & 98.74 & 71.35 & 72.21 & 81.78 & 88.41 & 60.48 & 66.30 & 47.64 & 62.95 & 46.19 & 0.00 & 49.53 & 72.60 & 77.71 & 83.22 & 59.53 & \textbf{64.91} & \textbf{76.04} \\
            \bottomrule
         \end{tabular}
    }
    \label{tab:Table.4}
    }
    \caption{(a)~Quantitative results of different pre-training datasets constructed by different unsupervised manners. SAMRS* represent the data generated by Oriented RepPoints \& SAMRS framework. (b)~Quantitative results of different pre-trained schemes on the iSAID dataset.}
\end{table*}

\subsection{Comparison to State-of-the-art Methods}
To demonstrate the superiority of our ALPS framework, we have compared it with the state-of-the-art automatic labeling pipeline SAMRS. However, the SAMRS adopts the ground truth detection annotations as input, our method is achieved in an unsupervised manner. Therefore, to achieve a fair comparison, we have achieved SAMRS framework in an unsupervised manner to construct a comparative dataset SAMRS* to demonstrate that our ALPS framework can generate a pre-training dataset that is more helpful for downstream segmentation tasks under same experimental configuration. 
Specifically, we have utilized the Oriented RepPoints \cite{oriented} based on mmrotate \cite{mmrotate}, loaded pre-trained weights, to provide object detection boxes for SAMRS to achieve an unsupervised manner. 
Moreover, we select UperNet and Segformer as the segmentation models and conduct experiments based on mmsegmentation with the same configuration to ensure the fairness and reliability of the comparison experiments.
The quantitative results are shown in Table \ref{tab:Table.6}, we can observe that the results pre-trained on our SAMRS-PL outperform all competitors in mIoU and mAcc compared to the results of SAMRS*, e.g., UperNet: Overall: 58.78 vs. 64.65, Backbone: 53.30 vs.57.45, Segformer: Overall: 64.80 vs. 65.35, Backbone: 64.19 vs. 64.70, which proves the effectiveness of the pseudo-labels generated by our ALPS framework. Furthermore, we also visualize the qualitative results in Fig \ref{fig:samrs_comp}, and we can observe that limited by the generalization of the detection model, the segmentation results obtained by SAMRS* actually get worse than ours in some ships and planes images. Benefiting from feature clustering in our mask class association, our method can generate segmentation maps with better visual effects without additional supervision information.

Please be advised that although the individual performance improvements might appear marginal, the cumulative effect across varying RS segmentation tasks highlights the robustness and generalized enhancement provided by ALPS.

\subsection{Ablation Study}
\label{sec:5.4}
\subsubsection{Effectiveness of Mask Class Association:} To verify the effectiveness of our proposed mask class association, we constructed a comparative baseline solely implemented by the binary mask prediction~(Sec \ref{sec:3.2}). Since the UiMs predicted by vanilla SAM do not correspond to specific target classes, it is reasonable to consider assigning these UiMs as a unified class to obtained the binary masks. Therefore, as shown in Fig \ref{fig:binary}, we have generated a set of binary masks for the iSAID dataset and then pre-training using UperNet, comparing the results with UperNet pre-training on our iSAID-PL. The quantitative results are shown in Table \ref{tab:Table.4}, we can discern a substantial disparity between the method pre-training on the binary masks and the methods pre-training on our generated datasets, denoted by a notable difference in performance metrics, e.g., UperNet: mIoU: 10.01 vs. 58.61 vs. 57.47, Segformer: mIoU: 11.03 vs. 64.30 vs. 64.70. Specifically, in the process of the binary mask prediction, we aggregate the binary mask set filtered through the filtering gate $G(\cdot)$ by stacking, thus forming a complete binary mask. To prevent large background masks from overshadowing instances, we set a relatively large area size threshold to filter out larger masks, retaining only smaller instances as foreground, which causes the pre-trained method demonstrates segmentation performance solely on the background, vehicle, plane and harbor. Nevertheless, benefiting from the feature clustering of our mask class association, ALPS can effectively avoid the above issues and generate pseudo-labels that can be used for pre-training to improve the performance of downstream tasks.

\begin{table*}[!t]
    \centering
    \scriptsize
    \resizebox{0.5\textwidth}{!}{
    \subfloat{
        \begin{tabular}{c|cccc|cc} 
        \toprule[1.2pt]
            Methods & Type & Pre-trained & Fine-tuned & Step & mIoU & mAcc \\
            \hline
            \multicolumn{3}{c}{\textbf{Convolution Based Network}} \\
            \hline
            \multirow{4}{*}{UperNet} 
                & Backbone & SAMRS-PL & iSAID & 80k & 57.47 & 64.68 \\
                & Backbone & SAMRS-PL & iSAID & 320k & 61.77 & 70.95 \\
                & Backbone & SAMRS-PL & Potsdam & 80k & 71.03 & 84.95 \\
                & Backbone & SAMRS-PL & Potsdam & 320k & 71.31 & 86.72 \\
            \hline
            \multirow{4}{*}{Pspnet} 
                & Backbone & SAMRS-PL & iSAID & 80k & 55.17 & 68.41 \\
                & Backbone & SAMRS-PL & iSAID & 320k & 60.28 & 70.65 \\
                & Backbone & SAMRS-PL & Potsdam & 80k & 70.56 & 84.57 \\
                & Backbone & SAMRS-PL & Potsdam & 320k & 71.17 & 85.15 \\
            \hline
            \multirow{4}{*}{HRNet} 
                & Backbone & SAMRS-PL & iSAID & 80k & 55.27 & 62.10 \\
                & Backbone & SAMRS-PL & iSAID & 320k & 59.24 & 70.88 \\
                & Backbone & SAMRS-PL & Potsdam & 80k & 69.50 & 84.21 \\
                & Backbone & SAMRS-PL & Potsdam & 320k & 71.05 & 85.22 \\
            \hline
            \multirow{4}{*}{Deeplabv3plus} 
                & Backbone & SAMRS-PL & iSAID & 80k & 52.93 & 60.78 \\
                & Backbone & SAMRS-PL & iSAID & 320k & 58.05 & 70.61 \\
                & Backbone & SAMRS-PL & Potsdam & 80k & 70.17 & 84.41 \\
                & Backbone & SAMRS-PL & Potsdam & 320k & 71.09 & 84.93 \\
            \bottomrule
         \end{tabular}
    }
    }
    \resizebox{0.48\textwidth}{!}{
    \subfloat{
        \begin{tabular}{c|cccc|cc} 
        \toprule[1.2pt]
            Methods & Type & Pre-trained & Fine-tuned & Step & mIoU & mAcc \\
            \hline
            \multicolumn{3}{c}{\textbf{Transformer Based Network}} \\
            \hline
            \multirow{4}{*}{ViT} 
                & Backbone & SAMRS-PL & iSAID & 160k & 61.22 & 72.16 \\
                & Backbone & SAMRS-PL & iSAID & 320k & 61.50 & 72.44 \\
                & Backbone & SAMRS-PL & Potsdam & 160k & 70.37 & 84.34 \\
                & Backbone & SAMRS-PL & Potsdam & 320k & 70.53 & 84.90 \\
            \hline
            \multirow{4}{*}{Swin} 
                & Backbone & SAMRS-PL & iSAID & 160k & 64.79 & 75.75 \\
                & Backbone & SAMRS-PL & iSAID & 320k & 64.95 & 75.97 \\
                & Backbone & SAMRS-PL & Potsdam & 160k & 71.72 & 85.49 \\
                & Backbone & SAMRS-PL & Potsdam & 320k & 72.19 & 85.70 \\
            \hline
            \multirow{4}{*}{Segmenter} 
                & Backbone & SAMRS-PL & iSAID & 160k & 57.37 & 66.28 \\
                & Backbone & SAMRS-PL & iSAID & 320k & 57.42 & 66.33 \\
                & Backbone & SAMRS-PL & Potsdam & 160k & 70.68 & 84.72 \\
                & Backbone & SAMRS-PL & Potsdam & 320k & 70.82 & 84.74 \\
            \hline
            \multirow{4}{*}{Segformer} 
                & Backbone & SAMRS-PL & iSAID & 160k & 64.70 & 75.04 \\
                & Backbone & SAMRS-PL & iSAID & 320k & 64.81 & 75.76 \\
                & Backbone & SAMRS-PL & Potsdam & 160k & 71.74 & 85.52\\
                & Backbone & SAMRS-PL & Potsdam & 320k & 71.86 & 85.99 \\
            \bottomrule
         \end{tabular}
    }
    }
    \caption{Quantitative results of larger pre-trained steps on the differernt datasets.}
    \vspace{-3mm}
    \label{tab:table.5}
\end{table*}

\subsubsection{Effectiveness of Larger Pre-training data:} To validate the larger pre-training data can improve the performance of fine-tuning, we have utilized the iSAID-PL \& SAMRS-PL into a Mixed-PL pre-training manner. Specifically, since the inconsistent number of categories in the two datasets, cannot be directly used for mixed training, we considerablely load the backbone weight obtained from the iSAID-PL to pre-train on the SAMRS-PL, with either 80k or 160k steps.
As shown in Table~\ref{tab:Table.4}, using the larger pre-trained data can effectively enhance the performance of UperNet and Segformer on iSAID dataset. We attribute this significant improvement to the data enhancement effect on the model backbone, which is achieved by the richer semantic features brought by an increased number of RS images.


\subsubsection{Effectiveness of Larger Pre-training Steps} To ensure the potential effectiveness of the pseudo-labels generated by our ALPS framework, we have pre-trained previous eight methods with larger pre-training steps~(e.g. 320k) on our SAMRS-PL dataset and fine-tuned on two validation datasets. The quantitative results are shown in Table. \ref{tab:table.5}. We can observe that the performance improves significantly when larger pre-training steps are employed. This enhanced performance surpasses previous results obtained by 80k or 160k steps when only the backbone weights of the pre-trained model are utilized for fine-tuning. Moreover, after 320k pre-training on SAMRS-PL dataset, a substantial enhancement in performance has been observed for the majority of convolution-based methods. However, the improvement is less pronounced for most transformer-based methods. We attribute this discrepancy is due to the inherent complexity of the self-attention mechanism, which may require a larger number of training steps to converge effectively.

\subsubsection{Area Size Threshold Selection}
During the dataset construction phase, we did not use a filtering gate to filter out overly large background images, which might have caused the SAM to treat all backgrounds other than the foreground target as a whole. However, SAM itself may have certain detection omissions, resulting in inconsistencies such as some targets, like vehicles, being classified as foreground while others are classified as background. This could potentially affect the effectiveness of pre-training. Therefore, we designed a filtering gate to filter out larger masks by a preset area size threshold. To select the appropriate area size threshold, we have chosen five different area size thresholds of 0.1, 0.3, 0.5, 0.7, and 1.0 to construct the dataset for pre-training and fine-tuning tasks. The results of our fine-tuning on Potsdam dataset are shown in the Table \ref{tab:Table.7}:

\begin{table}[!t]
    \centering
    \begin{tabular}{c|cc|cc}
    \toprule
        Area Size Threshold & Backbone & Step & mIoU & mAcc \\
        \midrule
        0.1 & UperNet & 80K & 70.17 & 83.73 \\
        0.3 & UperNet & 80K & \textbf{71.03} & \textbf{84.95} \\
        0.5 & UperNet & 80K & 65.22 & 76.90 \\
        0.7 & UperNet & 80K & 47.81 & 57.00 \\
        1.0 & UperNet & 80K & 42.99 & 54.35 \\
    \bottomrule
    \end{tabular}
    \caption{The ablation study of different area size thresholds.}
    \vspace{-5mm}
    \label{tab:Table.7}
\end{table}

From the above results, we can observe that when the area size threshold is 0.3, it obtains the best performance. The reason for poorer performance at other area size thresholds is analyzed as follows: 0.1 may filter out some larger main targets, such as large aircraft; 0.5 might retain some large background areas, for instance, SAM sometimes divides roads into sections, which could result in these road backgrounds being retained; 0.7 and 1.0 tend to retain larger backgrounds, thus affecting pre-training performance. Therefore, we selected an area size threshold of 0.3 for data generation in our specific experiments.

\section{Generalization on Medical Domain: an External Validation Study}
\begin{figure*}[!t]
    \centering 
    \includegraphics[width=1\linewidth]{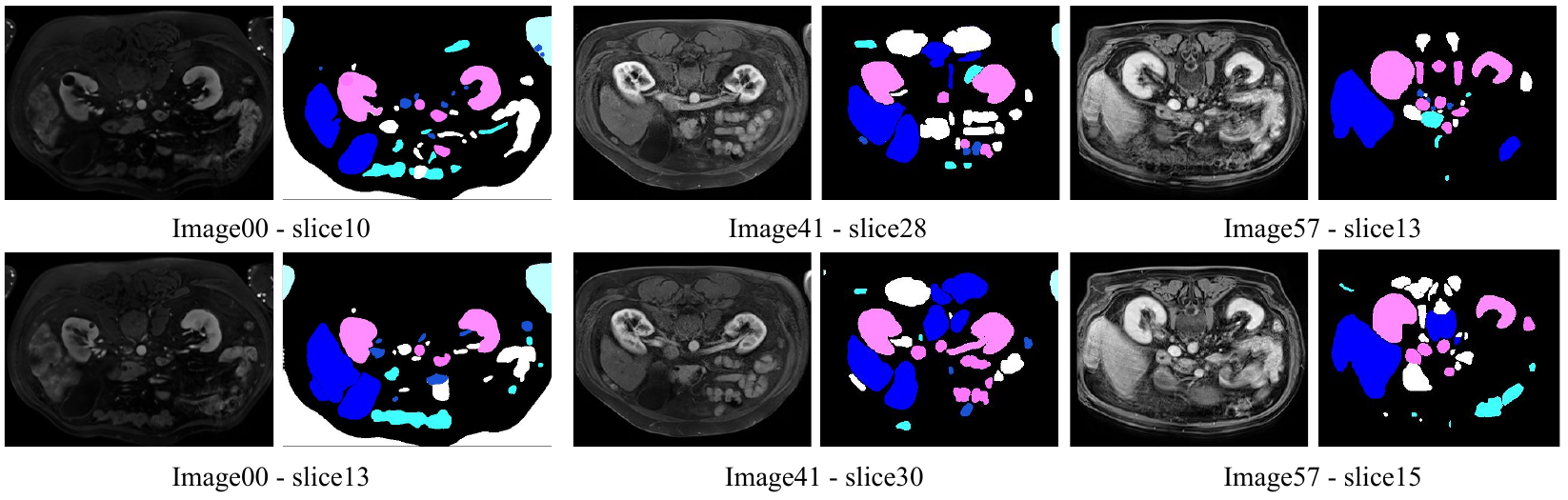}
    \caption{Some examples of Pseudo-Labeled results on ATLAS2023 dataset.} 
    \label{fig:med}
\end{figure*}

To further validate the generalization of ALPS on medical domain, we have utilized the ATLAS2023 dataset \cite{atlas}, a resource within the medical segmentation domain, to evaluate the efficacy of our ALPS for unsupervised auto-labeling of medical images. The ATLAS2023 dataset, designed for the automatic segmentation of tumors and livers, comprises T1 CE-MRI liver scans from 90 patients with unresectable liver cancer and 90 corresponding liver and liver tumor segmentation masks. These are divided into training and testing cohorts, consisting of 60 and 30 patients respectively. Given that the CE-MRI images (utilizing a gadolinium contrast agent) from the ATLAS dataset were acquired on five Siemens 3T and 1T MRI machines, this 3D MRI data is not directly applied to the vanilla SAM. To address this, we initially design a script to convert 3D MRI data into 2D PNG format, prior to implementing our framework for automatic annotation. The qualitative results are shown in Fig \ref{fig:med}. From the visualization results, we can observe that our proposed framework, designed for 2D slices of MRI data, demonstrates effective segmentation of liver and kidney. Furthermore, the color mapping within the framework accurately corresponds to the designated colors (pink for kidney and blue for liver) across multiple images. Due to the weak texture feature information of images, there may be unrelated instances that have been assigned to error labels. However, the label assignment for primary instances also maintains higher accuracy.

\begin{table}[!t]
    \centering
    \resizebox{1\linewidth}{!}{
    \begin{tabular}{cc|ccc|ccc}
    \toprule
       \multirow{2}{*}{ Methods } & \multirow{2}{*}{Type} & \multicolumn{3}{|c|}{ DSC (\%) } & \multicolumn{3}{c}{ mIoU (\%) } \\
       \cline { 3 - 8 } & & Liver & Tumor & Mean & Liver & Tumor & Mean \\
        \hline 
        \multirow{3}{*}{DoDNet} 
            & baseline & 79.54 & 24.03 & 51.79 & 75.31 & 19.3 & 47.31 \\
            & ATLAS-PL & 80.23 & 25.32 & 52.78 & 76.09 & 20.23 & 48.16 \\
            &  & $\textcolor{blue}{\textbf{+0.87\%}}$ & $\textcolor{blue}{\textbf{+5.37\%}}$ & $\textcolor{blue}{\textbf{+1.91\%}}$ & $\textcolor{blue}{\textbf{+1.03\%}}$ & $\textcolor{blue}{\textbf{+4.78\%}}$ & $\textcolor{blue}{\textbf{+1.80\%}}$ \\
        \hline
        \multirow{3}{*}{MED3D}
            & baseline & 73.12 & 19.55 & 46.34 & 65.30 & 16.82 & 41.06 \\
            & ATLAS-PL & 74.32 & 21.14 & 47.73 & 65.99 & 17.85 & 41.92 \\
            &  & $\textcolor{blue}{\textbf{+1.63\%}}$ & $\textcolor{blue}{\textbf{+8.12\%}}$ & $\textcolor{blue}{\textbf{+3.01\%}}$ & $\textcolor{blue}{\textbf{+1.07\%}}$ & $\textcolor{blue}{\textbf{+6.12\%}}$ & $\textcolor{blue}{\textbf{+2.09\%}}$ \\
        \hline
        \multirow{3}{*}{SAM-Med2D}
            & baseline & 80.02 & 51.28 & 65.65 & 77.34 & 48.88 & 63.11 \\
            & ATLAS-PL & 80.52 & 52.03 & 66.28 & 78.01 & 49.44 & 63.73 \\
            &  & $\textcolor{blue}{\textbf{+0.63\%}}$ & $\textcolor{blue}{\textbf{+1.46\%}}$ & $\textcolor{blue}{\textbf{+0.96\%}}$ & $\textcolor{blue}{\textbf{+0.87\%}}$ & $\textcolor{blue}{\textbf{+1.15\%}}$ & $\textcolor{blue}{\textbf{+0.98\%}}$ \\
    \bottomrule
    \end{tabular}
    }
    \caption{The ablation study of different area size thresholds.}
    \label{tab:Table.8}
\end{table}

To validate the pseudo-label ATLAS 2023 dataset (ATLAS-PL) we constructed, we selected mainstream methods for pre-training and fine-tuning, respectively. Specifically, we separately chose DoDNet \cite{Dodnet}, MED3D \cite{Med3d}, and SAM-Med2D \cite{Sam-med2d} for experimental validation. The experimental results are shown in Table \ref{tab:Table.8}, we can observe that after pre-training on our constructed ATLAS-PL, the performance of the above three methods has been significantly improved in both DSC and mIoU metrics. Specifically, for DoDNet, the DSC for Tumor increased by 5.37\% and the mIoU by 4.78\%. For MED3D, the DSC for Liver increased by 1.63\%, and the mIoU by 1.07\%. For SAM-Med2D, the increases in DSC and mIoU on average are 0.96\% and 0.98\%, respectively. We can attribute this significant improvement to our proposed ALPS framework, which can fully utilize the ability of vanilla SAM and generalization of online K-means to predict pseudo-label without medical prior for later pre-traning phrase.

\section{Limitation and Consideration}
In this study, our ALPS based vanilla SAM has been leveraged for data annotation tasks, demonstrating both time and cost efficiencies in processing extensive RS data. While our ALPS framework has successfully facilitated auto-labeling across a variety of segmentation tasks, it is imperative to acknowledge certain limitations that warrant further investigation: 

(1) \emph{Variability in Texture Features:} Instances within the same category may exhibit significant textural variation, potentially leading to diverse class assignments during feature clustering. Future research could explore the utilization of structured feature extraction to enhance clustering accuracy.

(2) \emph{Dependency on SAM Segmentation:} The reliance on SAM for pseudo-labels introduces variability, such as the segmentation of a single object into multiple masks or the amalgamation of multiple objects into a single mask. This inconsistency can affect mask class association and the overall quality of the generated dataset. Future directions might include integrating DINO \cite{dino} to provide targeted information as prompts to SAM, aiming to refine mask predictions. 

(3) \emph{Novelty and Significance:} ALPS leverages the pre-trained SAM, which is originally trained on human-annotated data. However, in ALPS, no additional manual annotations are required for RS images, thus enabling a semi-automated annotation process without additional human intervention. Further, though every component in the pipeline has been well explored, the application of clustering algorithms in conjunction with SAM for RS images and the novel mechanism for aligning pseudo-labels represents a noteworthy adaptation and refinement for RS contexts. Finally, the individual performance improvements might appear marginal, however the cumulative effect across varying RS segmentation tasks highlights the robustness and generalized enhancement provided by ALPS.

\section{Conclusion}
In conclusion, this paper introduces ALPS, an auto-labeling framework that capitalizes on the unmatched feature processing abilities of the vanilla Segment Anything Model (SAM) to efficiently annotate large-scale remote sensing (RS) datasets without additional prompts. Different from previous SAM extensions requiring additional detection prompts, ALPS adopts an innovative approach by extracting and clustering mask features to accurately predict PCL for each segmentation mask. To ascertain the efficacy of ALPS, we meticulously constructed two pseudo-labeled RS datasets, iSAID-PL and SAMRS-PL, without resorting to external annotations. Comprehensive experiments conducted on these datasets have underscored ALPS's considerable impact on enhancing downstream task performances, establishing a new benchmark in the field. Furthermore, through rigorous ablation studies, the crucial role of mask class association and the incremental benefits of extending pre-training steps have been affirmed, attesting to the robustness and indispensability of each component within our framework. Marking a pioneering effort to employ SAM devoid of any prompts for auto-labeling, this study lays the groundwork for innovative feature clustering applications combined with computer vision foundation models. Our future work aim to explore the broader implications of our approach across different domains while striving to refine the precision of auto-labeling techniques further.

\bibliography{aaai24}
\clearpage

\end{document}